\documentclass{bmvc2k}

\usepackage{pgfplots}
\usepackage[utf8]{inputenc}

\usepackage{epsfig}
\usepackage{graphicx}
\usepackage{amsmath}
\usepackage{amssymb}

\usepackage{color}
\definecolor{fsublue}{RGB}{62,106,190}
\definecolor{craneblue}{RGB}{4,6,76}

\usepackage{tabularx}
\usepackage{footmisc}
\usepackage{booktabs}
\usepackage{bm}

\newcommand\myparagraph[1]{
\vspace{-4pt}
\paragraph{#1}
}
\newcommand\sectionname{Sect.}
\newcommand\equationname{Eq.}

\def\eg{\emph{e.g}\bmvaOneDot}

\def\etal{\emph{et al}\bmvaOneDot}
\def\ie{\emph{i.e}\bmvaOneDot}

\newcommand{\thead}[1]{\textbf{\textsc{#1}}}
\newcolumntype{L}[1]{>{\raggedright\arraybackslash}m{#1}}
\newcolumntype{R}[1]{>{\raggedleft\arraybackslash}m{#1}}
\newcolumntype{C}[1]{>{\centering\arraybackslash}m{#1}} 

\newcommand\wscheme[1]{\thead{#1}\includegraphics[height=1.2ex]{figures/weights/#1-sym.pdf}}

\newcommand\predictor{f}
\newcommand\inputspace{\mathcal{X}}
\newcommand\outputspace{\mathcal{Y}}
\newcommand\inputsingle{\bm{x}}
\newcommand\outputsingle{y}

\newcommand\dataset{\mathcal{D}}
\newcommand\noe{n}
\newcommand\regul{\mathcal{R}}
\newcommand\loss{\mathcal{L}}
\newcommand\params{\bm{\theta}}
\newcommand\thetime{t}

\DeclareMathOperator\argmin{\text{argmin}}

\graphicspath{{figures/}}

\title{Impatient DNNs -- Deep Neural Networks with Dynamic Time Budgets}

\addauthor{Manuel Amthor}{manuel.amthor@uni-jena.de}{1}
\addauthor{Erik Rodner}{erik.rodner@uni-jena.de}{1}
\addauthor{Joachim Denzler}{joachim.denzler@uni-jena.de}{1}

\addinstitution{
    Computer Vision Group\\
    Friedrich Schiller University Jena \\
    Germany\\
    \small\textcolor{bmv@sectioncolor}{\bmvaUrl{www.inf-cv.uni-jena.de}}
}

\runninghead{Manuel Amthor, Erik Rodner, and Joachim Denzler}{Impatient DNNs}
\begin{document}

\maketitle

\begin{abstract}
    We propose Impatient Deep Neural Networks (DNNs) which deal with
    dynamic time budgets during application. They allow for individual budgets given 
    a priori for each test example and for anytime prediction, \ie a possible interruption at multiple
    stages during inference while still providing output estimates. 
    Our approach can therefore tackle the computational costs and energy demands 
    of DNNs in an adaptive manner, a property essential for real-time applications.

    Our Impatient DNNs are based on a new general framework of learning dynamic budget predictors
    using risk minimization, which can be applied to current DNN architectures
    by adding early prediction and additional loss layers. A key aspect of our method is that
    all of the intermediate predictors are learned jointly.
    In experiments, we evaluate our approach for different budget distributions,
    architectures, and datasets. Our results show a significant gain in expected accuracy compared
    to common baselines.
\end{abstract}

\section{Introduction}
\label{sec:intro}

Deep and especially convolutional neural networks are the current base for the majority
of state-of-the-art approaches in vision. Their ability to learn very
effective representations of visual data has led to several breakthroughs in important applications, such
as scene understanding for autonomous driving~\cite{cordts2016cityscapes}, object detection~\cite{girshick2014rich},
and robotics~\cite{finn2016deepspatial}.
The main obstacle for their application is still the computational cost
during prediction for a new test image. Many previous works have focused on speeding up DNN inference
in general achieving constant speed-ups for a certain loss in prediction accuracy~\cite{jaderberg2014speeding,lebedev2015fast}.

\begin{figure}
    \centering
    \includegraphics[width=0.99\linewidth]{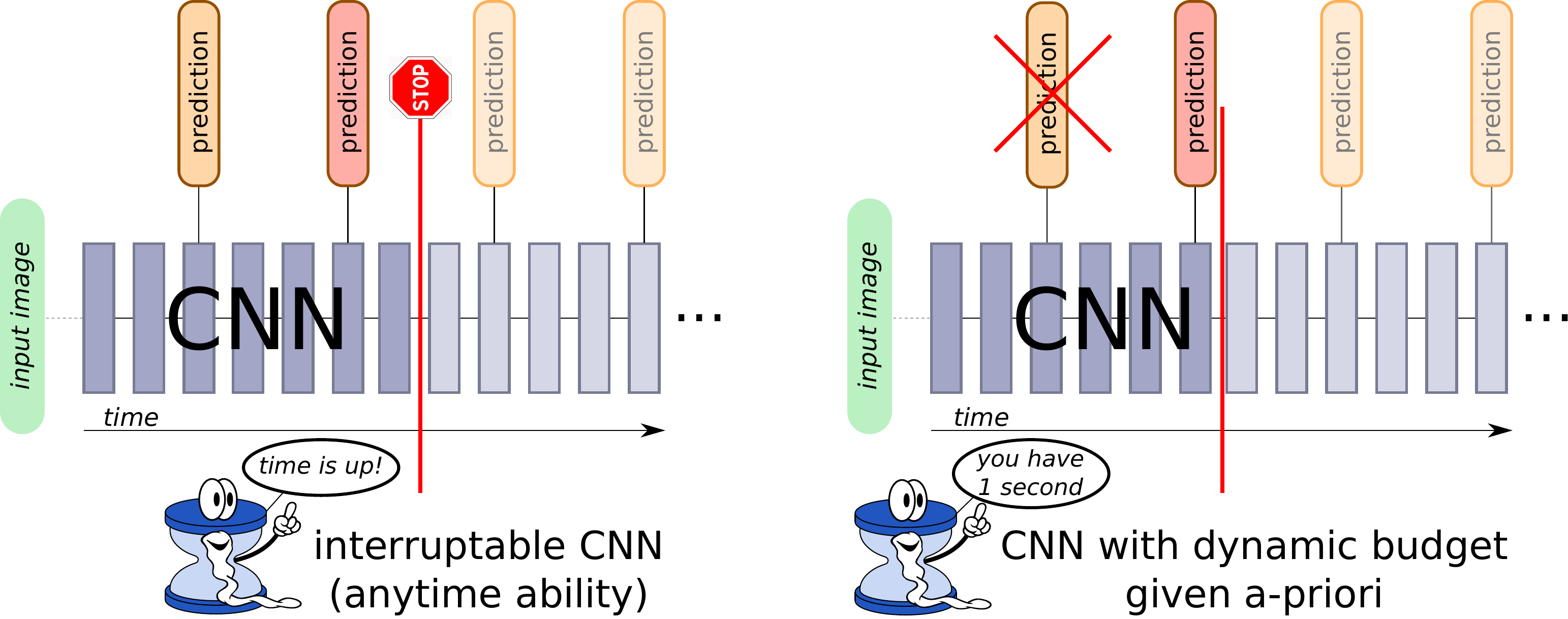}
    \caption{Illustration of convolutional neural network prediction in dynamic budget scenarios: (left) prediction can be interrupted at \textbf{any time} or (right) the budget is given \textbf{before} each prediction.}
    \label{fig:teaser}
\end{figure}

In contrast, we focus on inference with dynamic time budgets. Our networks provide a series of
predictions with increasing computational cost and accuracy. This allows for (1) dynamic interruption of the prediction
in time-critical applications (anytime ability, \figurename~\ref{fig:teaser} left), or for (2) predictions with a dynamic time budget individually given for each test image a-priori (\figurename~\ref{fig:teaser} right).
Dynamic budget approaches can for example deal with varying energy resources, a property especially useful for real-time visual inference
in robotics~\cite{levine2015end}. Furthermore, early predictions allow for immediate action selection in reinforcement learning scenarios~\cite{silver2013learning}.

The main idea of our approach is to formulate the learning of dynamic budget predictors as a generalized risk minimization that involves
the distribution of budgets provided for the application. The distribution of possible budgets has been either previously neglected
or assumed to be uniform~\cite{karayev2014anytime}. However, we show that such an easily available prior information
can significantly help to improve the expected accuracy.

Our formulation leads to a straight-forward modification of convolutional neural network (CNN)
architectures and their training. In particular, we add several early prediction and loss layers along the standard processing pipeline of a DNN (\figurename~\ref{fig:teaser} and \figurename~\ref{fig:architecture}). 
According to our risk minimization framework for dynamic budget predictors, all of these layers need to be learned jointly with a weighted combination derived from a time-budget distribution. Whereas this strategy is directly related to DNN learning strategies, such as deep supervision~\cite{wang2015training} and
inception architectures~\cite{szegedy2015going}, we demonstrate its usefulness for adapting to varying resources during testing.

The paper is structured as follows. After discussing related work, we define dynamic budget 
predictors and derive a new learning framework based on risk minimization with budget distributions 
(\sectionname~\ref{sec:anytimepred}). Our framework can be directly applied to deep and especially convolutional neural networks
as described in \sectionname~\ref{sec:impatientcnns}. Experiments in \sectionname~\ref{sec:exp} show
the advantages of our approach for different architectures, datasets, and budget distributions.

    \myparagraph{Related work on anytime prediction}
    The work of Karayev~\etal~\cite{karayev2014anytime} presented an approach that iteratively 
    and dynamically selects feature representations to maximize the area above an entropy vs. cost curve.
    Our approach however focuses on a static order of predictors and is able to incorporate
    budget distributions expected for the application.
    Fröhlich~\etal~\cite{Froehlich12:ATG} proposed a semantic segmentation approach
    with anytime classification capability. Their method is based on random decision forests
    learned in a layer-wise fashion. 
    Xu~\etal~\cite{xu2013anytime} considers anytime classification with unknown budgets by
    combining a cost-sensitive support vector machine with feature learning. 
    Similar to \cite{Froehlich12:ATG}, their predictors are learned in a greedy fashion and not learned jointly as in our case.
    Learning all of the predictors with shared parameters jointly allows us to share computations while
    directly optimizing with respect to expected accuracy during training.
    The paper of \cite{xu2012greedy} presents an algorithm for learning tree ensembles
    with a constrained time budget available during training. In our case, the whole distribution
    budgets is given during training.

    \myparagraph{Related work on deep supervision and DNNs with multiple losses}
    There are multiple methods that use a similar architecture of deep neural networks
    than ours characterized by multiple loss layers and joint training of them.
    For example, \cite{wang2015training} refers to such a training strategy as 
    ``deep supervision'' and shows that it allows for training deeper networks in a robust fashion.
    A very similar technique has been used in \cite{guo2016locally} for improved scene recognition.
    Furthermore, multiple loss layers are often used for multi-task learning, where the goal is
    to jointly predict various outputs~\cite{zhang2016learning}.

    In contrast to these works, our paper focuses on the impact of such an architecture
    on the ability of DNNs to deal with dynamic time budgets during inference. Furthermore,
    we show that such an architectural design can be directly derived from a very general
    risk minimization framework for predictors with dynamic budgets.

    \myparagraph{Related work on speeding up convolutional neural networks}

    There are multiple works that focus on speeding up DNNs and the special case of 
    convolutional neural networks (CNNs). Applied and adapted techniques
    range from low-rank approximations~\cite{jaderberg2014speeding,girshick2014rich,denton2014exploiting} to
    FFT computations of the involved convolutions~\cite{mathieu2013fast}.
    The Fast R-CNN method of \cite{girshick2014rich} speeds up fully-connected layers by simple SVD approximation.
    Similar techniques have been presented by \cite{denton2014exploiting} and \cite{jaderberg2014speeding}.
    The paper of \cite{He14:CNN} provides an empirical study of the effects of CNN architectural design choices
    on the computation time and the achieved recognition performance.
    A straightforward technique to speed up convolutions with large filter sizes uses Fast Fourier Transforms as
    studied by \cite{mathieu2013fast}. Furthermore, efficient filtering techniques, such as the Winograd transformation~\cite{lavin2015fast},
    are applicable as well.

    Our approach also tries to speed up inference of deep neural networks, \ie a forward pass. However,
    instead of approximating different operations performed in single layers, we achieve a significant speed-up by
    allowing the algorithm to deal with dynamic time budgets. Therefore, our research is orthogonal to the one briefly described 
    and combining them is straightforward.

\section{Learning Dynamic Budget Predictors}
\label{sec:anytimepred}

In this section, we derive a simple yet powerful learning scheme for dynamic budget predictors. Without
loss of generality, we focus on time budgets in the following.

\myparagraph{Specification of dynamic budgets}
    An important challenge for dynamic budget approaches is that the budget available for
    inference during testing is not known during training and for anytime scenarios even not known
    during inference itself. 
    For anytime tasks, we need to learn algorithms that can be interrupted at several time steps
    and balance the trade-off between calculating direct predictions of an output $\outputsingle$
    for an example $\inputsingle$ or performing calculating intermediate outputs that help later
    on for further refinements of the predictions.

    This trade-off is without any further specification, ill-posed. However, in many applications,
    we know something about the distribution $p(\thetime \;|\; \inputsingle, \outputsingle)$ 
    of time budgets $\thetime$ available to the algorithm for a given input-output pair $(\inputsingle, \outputsingle)$.
    In the following, we assume that this distribution is either given or can be modeled for
    an application. 
    
    \myparagraph{Risk minimization with budget distributions}
    In the following, 
    we develop a framework for learning dynamic budget predictors using risk minimization.
    We consider inference algorithms $\predictor$ that provide predictions $\outputsingle \in \outputspace$ 
    for input examples $\inputsingle \in \inputspace$
    at different times $\thetime \in \mathbb{R}$, \ie we have $\predictor: \inputspace \times \mathbb{R} \rightarrow \outputspace$.

    Learning the parameters $\params$ of $\predictor$ 
    is done by minimizing the following regularized risk:
    \begin{align}
        \label{eq:emprisk1}
        &\argmin_{\params}\, \int_{\thetime \in \mathbb{R}} \int_{\outputsingle \in \outputspace} \int_{\inputsingle \in \inputspace} 
        \loss( \predictor(\inputsingle, \thetime; \params), \outputsingle ) \cdot p(\inputsingle, \outputsingle, \thetime) \,\mathrm{d}\inputsingle\, \mathrm{d}\outputsingle\, \mathrm{d} \thetime + \regul(\params) \enspace,
    \end{align}
    with $\loss$ being a suitable loss function, $\regul(\params)$ being a regularization term,
    and $p(\inputsingle, \outputsingle, \thetime)$ being the joint distribution of
    an input-output pair $(\inputsingle, \outputsingle)$ and the available time $\thetime$.
    This formulation does not require any differentiation between a-priori given budget or anytime scenarios.

    We further assume that the time available is independent of the actual
    example and it's label. 
    This is a reasonable assumption, since the available time is in most
    applications just based on a limitation of hardware or data transfer resources.
    Since we are given a training set $\dataset = (\inputsingle_i, \outputsingle_i)_{i=1}^{\noe}$,
    learning is based on minimizing the empirical risk:
    \begin{align}
        \label{eq:emprisk2}
        &\argmin_{\params}\, \int_{\thetime \in \mathbb{R}} \sum\limits_{i=1}^{\noe} 
        \loss( \predictor(\inputsingle_i, \thetime; \params), \outputsingle_i ) \cdot p(\thetime) \mathrm{d} \thetime + \regul(\params) \enspace.
    \end{align}
    
    The predictor $\predictor$ is an algorithm performing a finite sequence of atomic operations. Therefore, 
    the prediction output will be only changing at discrete time steps $t_1, \ldots, t_K$:
    \begin{align}
        \forall \inputsingle \;\forall 1 \leq k < K\; \forall t_k \leq t < t_{k+1}:\quad& \predictor(\inputsingle, t; \params) = \predictor(\inputsingle, t_k; \params) \stackrel{\text{def.}}{=} \predictor_k(\inputsingle; \params_k),\\
        \forall \inputsingle \; \forall t \geq t_K:\quad& \predictor(\inputsingle, t; \params) = \predictor_K(\inputsingle; \params_K)\enspace.
    \end{align}
    Furthermore, before $t_1$, no output estimate is available. Since this leads to a constant additive term independent
    of $\params$, we can ignore this aspect in the following.
    In total, \equationname~\eqref{eq:emprisk2} simplifies as follows:
    \begin{align}
        \label{eq:emprisk3}
        &\argmin_{\params}\, \sum\limits_{k=1}^{K} w_k \cdot \left( \sum\limits_{i=1}^{\noe} 
    \loss( \predictor_k(\inputsingle_i; \params_k), \outputsingle_i ) \right) + \regul(\params) \enspace,
    \end{align}
    with weights $w_k = \int_{t_k}^{t_{k+1}} p(t) \mathrm{d}t$ for $1 \leq k < K$ and $w_K = \int_{t_K}^{\infty} p(t) \mathrm{d}t$. 
    As can be seen we have a simple learning objective, which is a weighted combination of the learning objectives of each of the
    individual predictors $\predictor_k$. If some of the parameters are shared between the predictors, which is the case for our approach presented in \sectionname~\ref{sec:impatientcnns}, 
    each term in the objective can not be optimized independently and joint optimization is necessary.
    Sharing parameters is essential for optimizing shared computations towards maximizing the expected accuracy of the complete model.
    
    The information about the time-budget distribution defines the weights of the loss terms in an intuitive manner:
    if there is a high probability of the time budget being between
    $\thetime_{k}$ and $\thetime_{k+1}$, the loss of $\predictor_k$ has a strong impact on the overall learning objective and
    the parameters $\params_k$ including the shared ones should be tuned towards reducing the loss of $\predictor_k$ rather
    than contributing significantly to other predictors. 

\section{Learning Impatient DNNs with Early Prediction Layers}
\label{sec:impatientcnns}

    \begin{figure}[tb]
        \centering
        \includegraphics[width=0.9\linewidth]{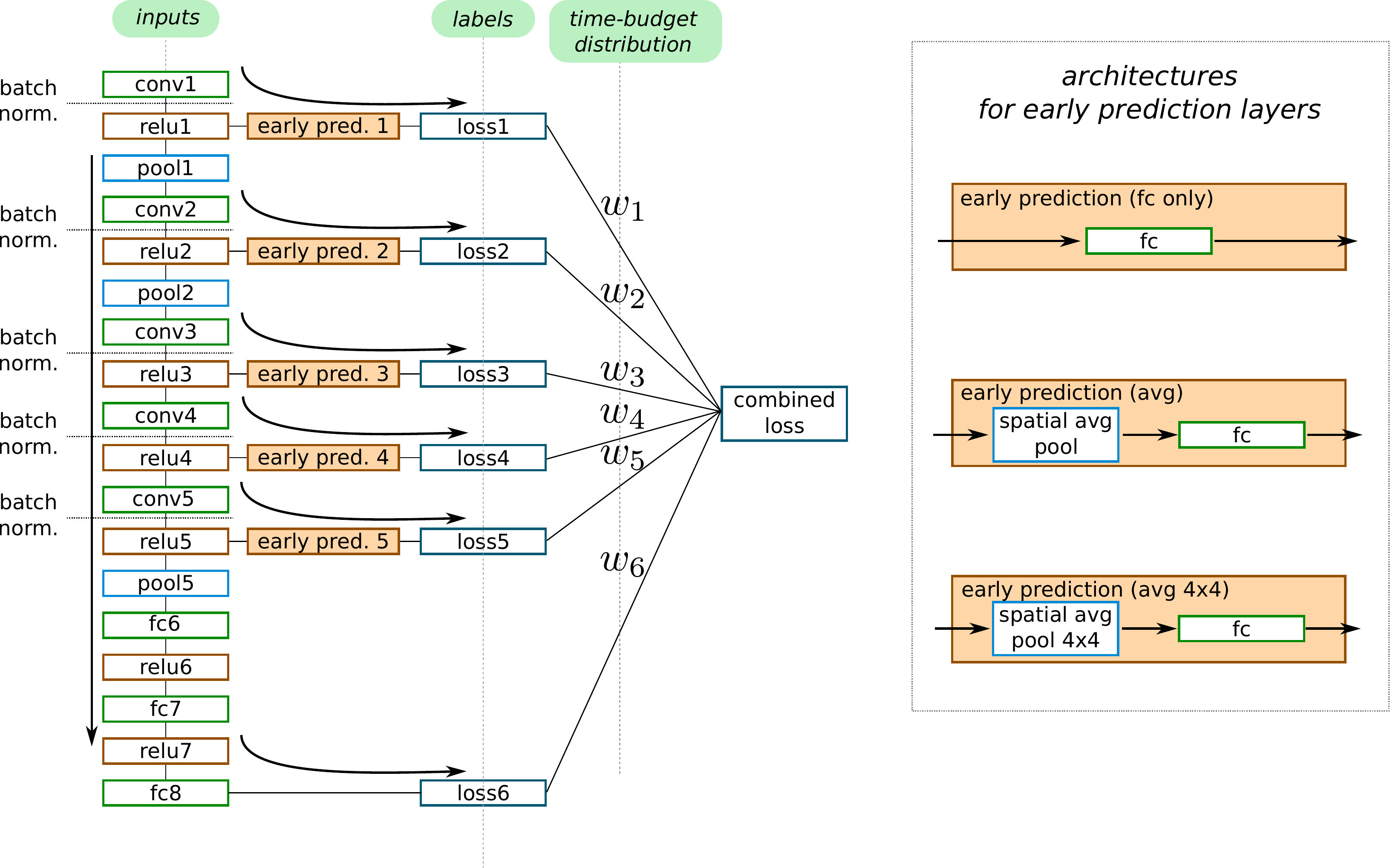}
        \caption{(Left) Modification of the AlexNet architecture for dynamic budgets and early predictions. (Right) Possible architectures
        for early prediction.}
        \label{fig:architecture}
    \end{figure}

    In this section, we show how a single deep neural network
    with additional prediction layers is well suited for providing a series
    of prediction models.

    \myparagraph{Early prediction layers}
    To obtain a series of predictions, we add
    $K$ additional layers to a common DNN architecture as 
    illustrated in \figurename~\ref{fig:architecture}.
    We refer to these layers as early prediction (EP) layers in the following.
    The output $\predictor_k(\inputsingle)$ of these layers has as many dimensions as $\outputsingle$.
    Already after the first layers, our approach is able to perform predictions
    with only a very few number of computational operations. 
    The layered architecture of a DNN has an important advantage, since 
    all $\predictor_k$ naturally share a large set of their parameters and 
    also a large number of computations. 
    Anytime approaches
    require a forward pass to go through all 
    early prediction layers that can be processed until interruption. 
    In case of non-parallel computation, the computational 
    overhead of the early prediction layers should therefore be reduced
    as much as possible. 
    
    The right part of \figurename~\ref{fig:architecture}
    shows different choices for EP layers we experimented
    with: (1) FC only, which is a simple single fully-connected (FC) layer followed by a softmax layer,
    (2) AVG, which performs average pooling across the spatial dimensions of previous layer before 
    a fully-connected layer, which leads to a significantly reduced number of parameters for the EP layers,
    and (3) AVG $4 \times 4$, which allows for preserving rough spatial information
    by performing average pooling in $4 \times 4 = 16$ uniformly-sized regions.
    
    \myparagraph{Learning with weighted losses}
    For learning, each of the EP layers is connected to a loss layer.
    The overall loss during training is exactly the weighted combination we derived in the
    previous section in \equationname~\eqref{eq:emprisk2}.

    In theory, training our Impatient DNNs does not require any further modifications
    and learning can be done with standard back-propagation and gradient-descent.
    However, we observed in experiments that batch normalization~\cite{ioffe2015batch} leads
    to a significantly more robust training and is even required to achieve convergence
    at all in most cases.

    \myparagraph{Weighting schemes}
    \begin{figure}[tb]
        \includegraphics[width=0.24\linewidth]{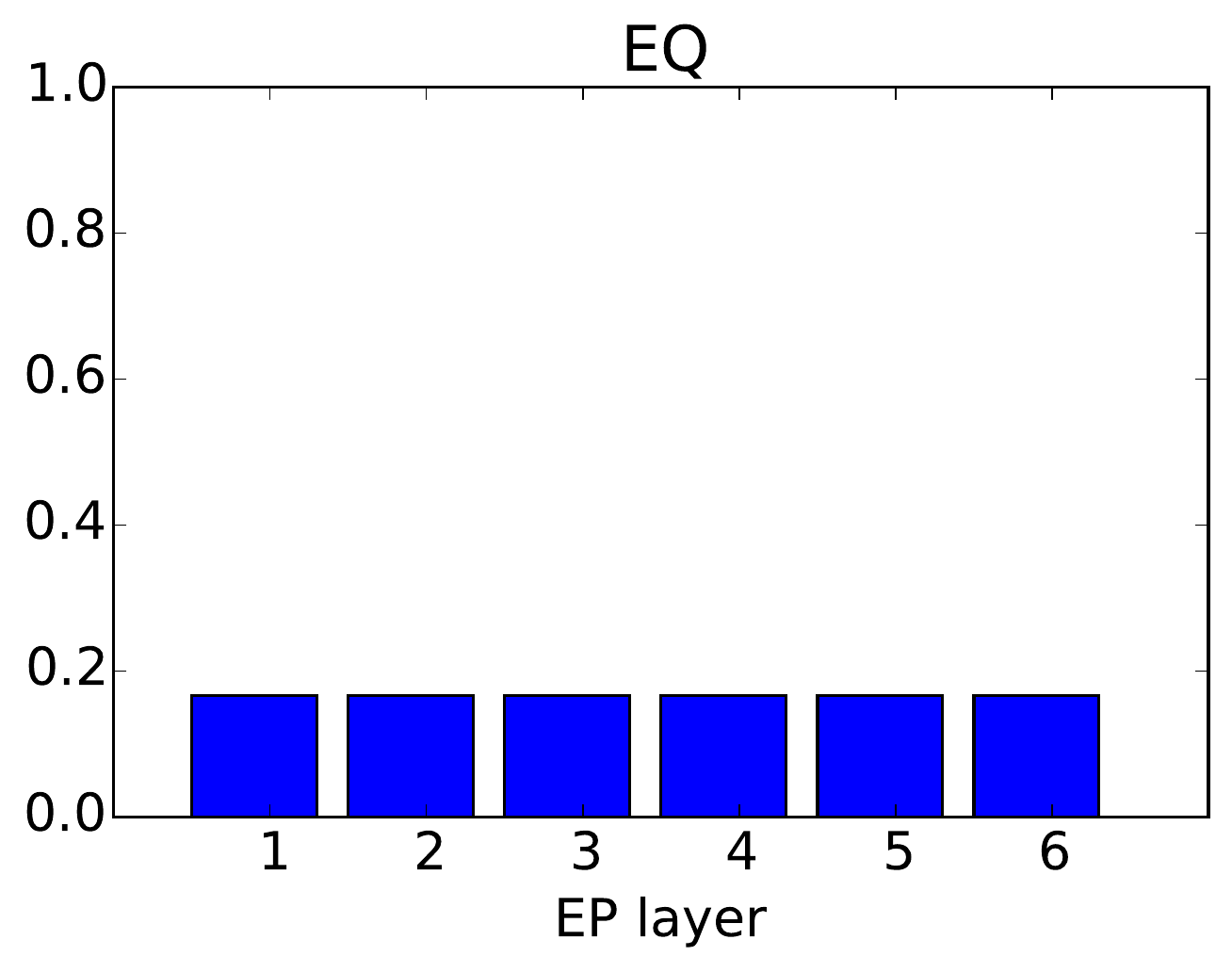}
        \includegraphics[width=0.24\linewidth]{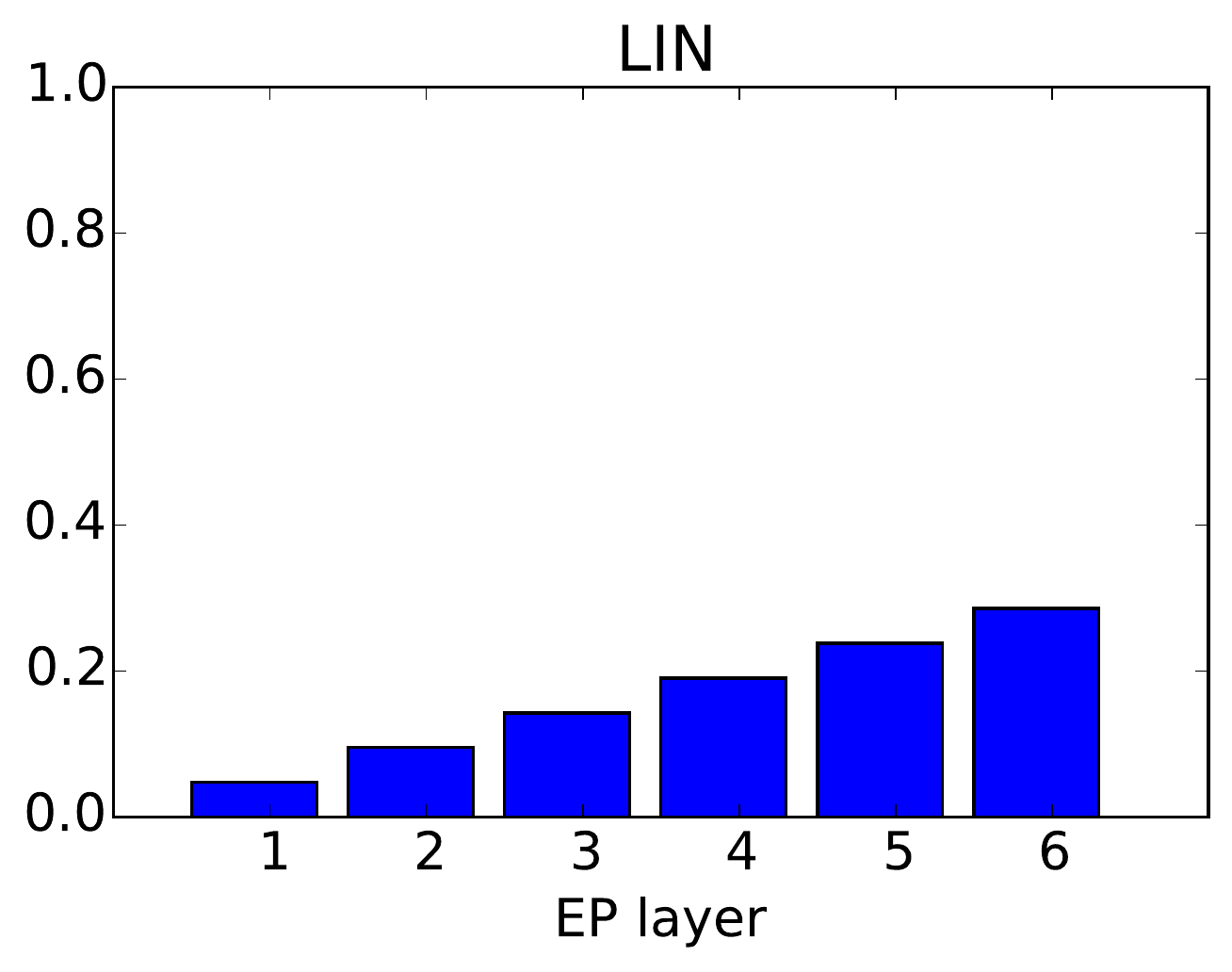}
        \includegraphics[width=0.24\linewidth]{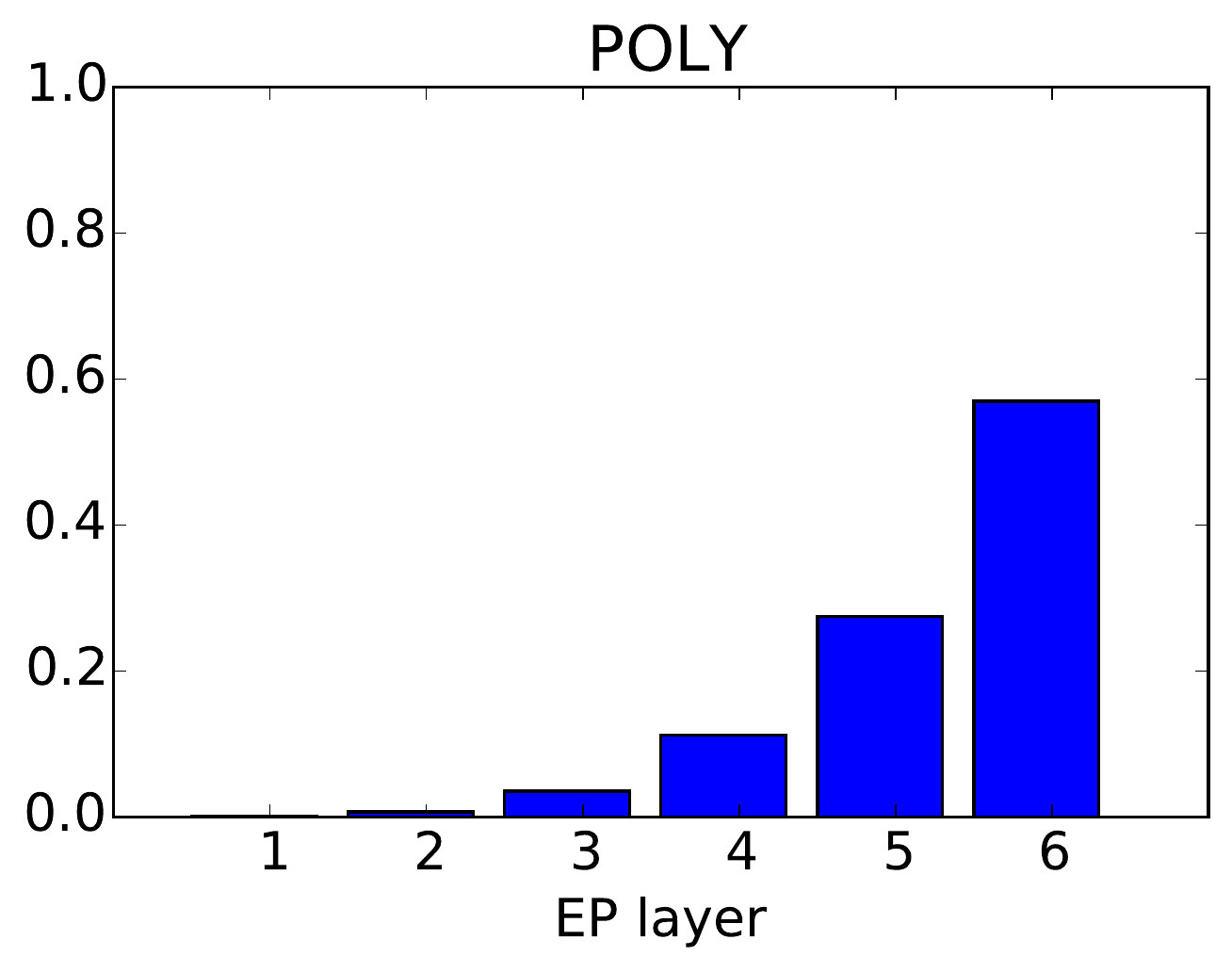}
        \includegraphics[width=0.24\linewidth]{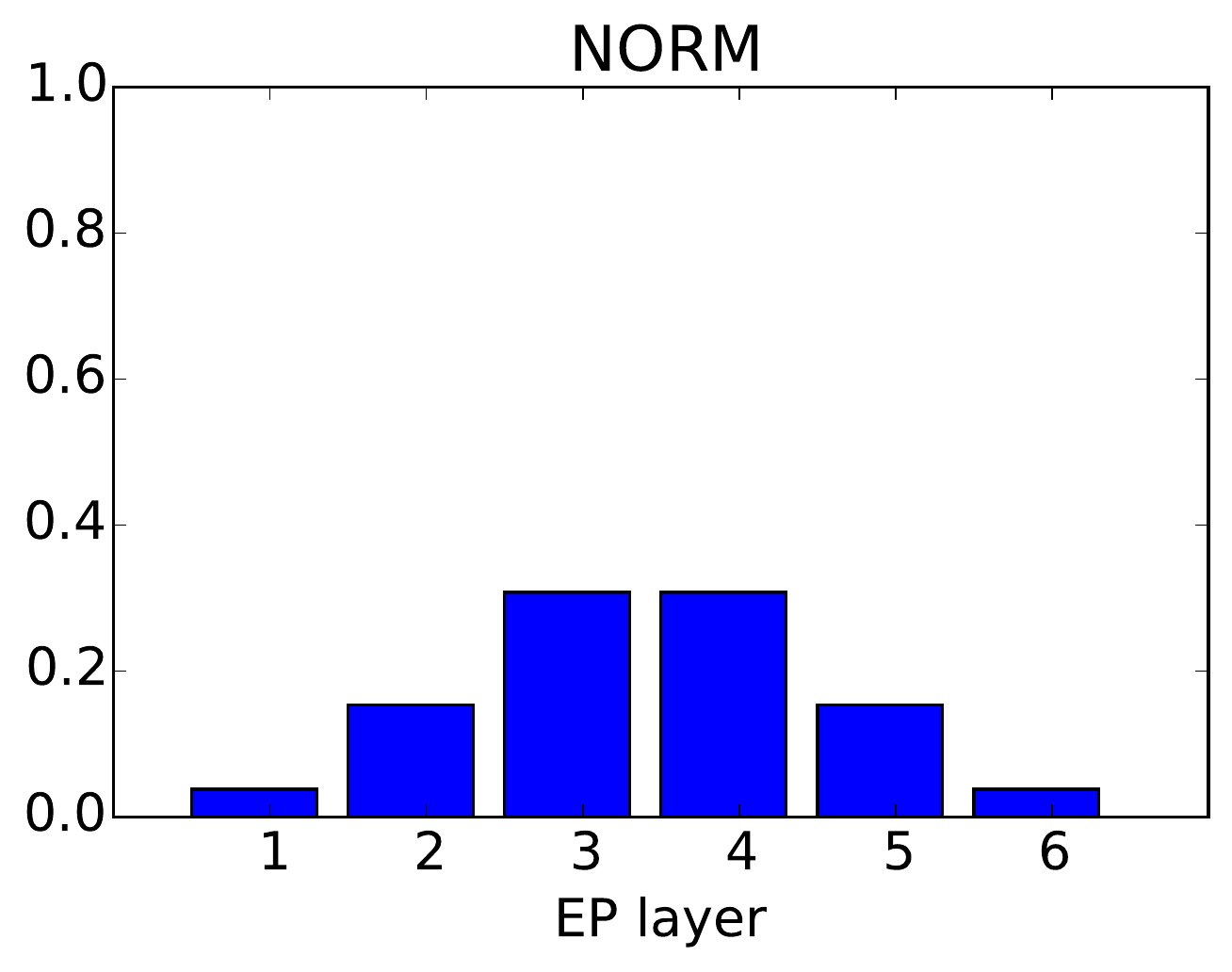}
        \caption{Types of time-budget distributions we consider in our paper.}
        \label{fig:budgettypes}
    \end{figure}
    In our experiments, we are interested in the effect of different time-budget distributions provided during learning.
    To simulate them, we consider the following schemes for early prediction layer weights $w_1, \ldots, w_K$:
    (STD) standard DNN training, \ie only the last prediction matters: $w_K = 1$ and $w_k = 0$ otherwise, 
    (EQ) uniform weights for uniform time-budget distributions: $w_k = \frac{1}{K}$, 
    (LIN) linearly increasing weights, \ie small time budgets are unlikely: $w_k \propto k$,
    (POLY) polynomially increasing weights: $w_k \propto k^\gamma$ with $\gamma > 1$,
    (ILIN, IPOLY) decreasing weights, \ie small time budgets are likely: $w_k = w_{K+1-k}'$ for weights $w_k'$ of the former schemes,
    and (NORM) small and large time budgets are rare and layers in the middle of the architecture are given a high weight: $w_k \propto \exp(- \beta \cdot (k-\frac{K+1}{2})^2)$ with $\beta = 0.34$.
    All of these schemes are simulating different budget specifications of an application.
    An illustration of several instances is given in \figurename~\ref{fig:budgettypes}.

\section{Experiments}
\label{sec:exp}
    
    In the following, we evaluate our approach with respect to different dynamic budget schemes and compare with
    standard DNN training and other relevant baselines.
    
    \myparagraph{Experimental setup and datasets}
    
    For evaluation, we conducted experiments on two object classification datasets.
    The \textbf{15-Scenes} \cite{lazebnik2006beyond} dataset comprises a total of 4,485 images covering categories from kitchen and living room to suburban and industrial.
    Each category contains between 200 and 400 images each, from which we took 100 images for training, as suggested by \cite{lazebnik2006beyond},
    and the remaining ones for testing.
    The training set is further divided into 90 images for actual training and 10 images for validation.
    The \textbf{MIT-67} \cite{quattoni2009recognizing} indoor scenes database is comprised of 67 categories.
    We follow the procedure of \cite{quattoni2009recognizing} and take 80 images for training and 20 for testing.
    Again, the training set is split in order to obtain a validation set of 8 images per class.

    Since our datasets are too small for DNN training from scratch, we perform fine-tuning of different models pre-trained on ImageNet, \eg\ AlexNet \cite{krizhevsky2012imagenet} and VGG19 \cite{simonyan2014very}. The positions of EP layers for AlexNet are given in
    \figurename~\ref{fig:architecture}. For VGG19, we add EP layers after each block of convolutional layers. Please note that the last ``early'' prediction layer is always the output layer of the original CNN architecture. 

    \myparagraph{Analysis of learning Impatient DNNs}

    In the following, we show that for learning Impatient DNNs care has to be taken to ensure convergence.
    For example, an adequate learning rate has to be determined to ensure convergence of the network while avoiding saturation at low accuracy.
    This becomes much more important when dealing with losses of multiple branches, since the gradients at shared layers accumulate
    leading to the network training being more fragile.
    Especially in the case of deeper network architectures, \eg\ VGG, we observed that convergence can not be achieved at all without proper normalization.

    \begin{figure}[t]
        \centering
        \includegraphics[height=0.37\linewidth]{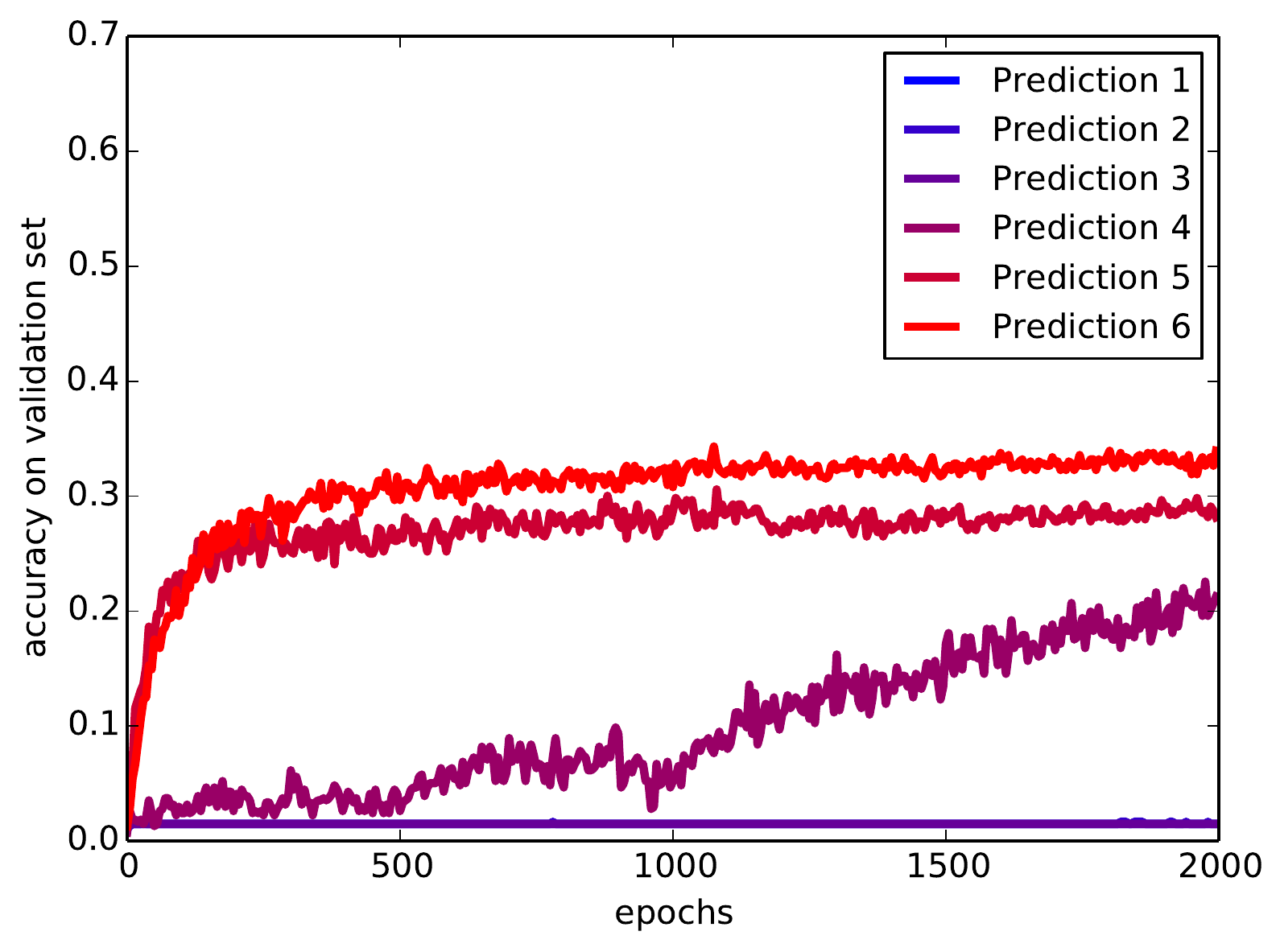} \hfil %
        \includegraphics[height=0.37\linewidth]{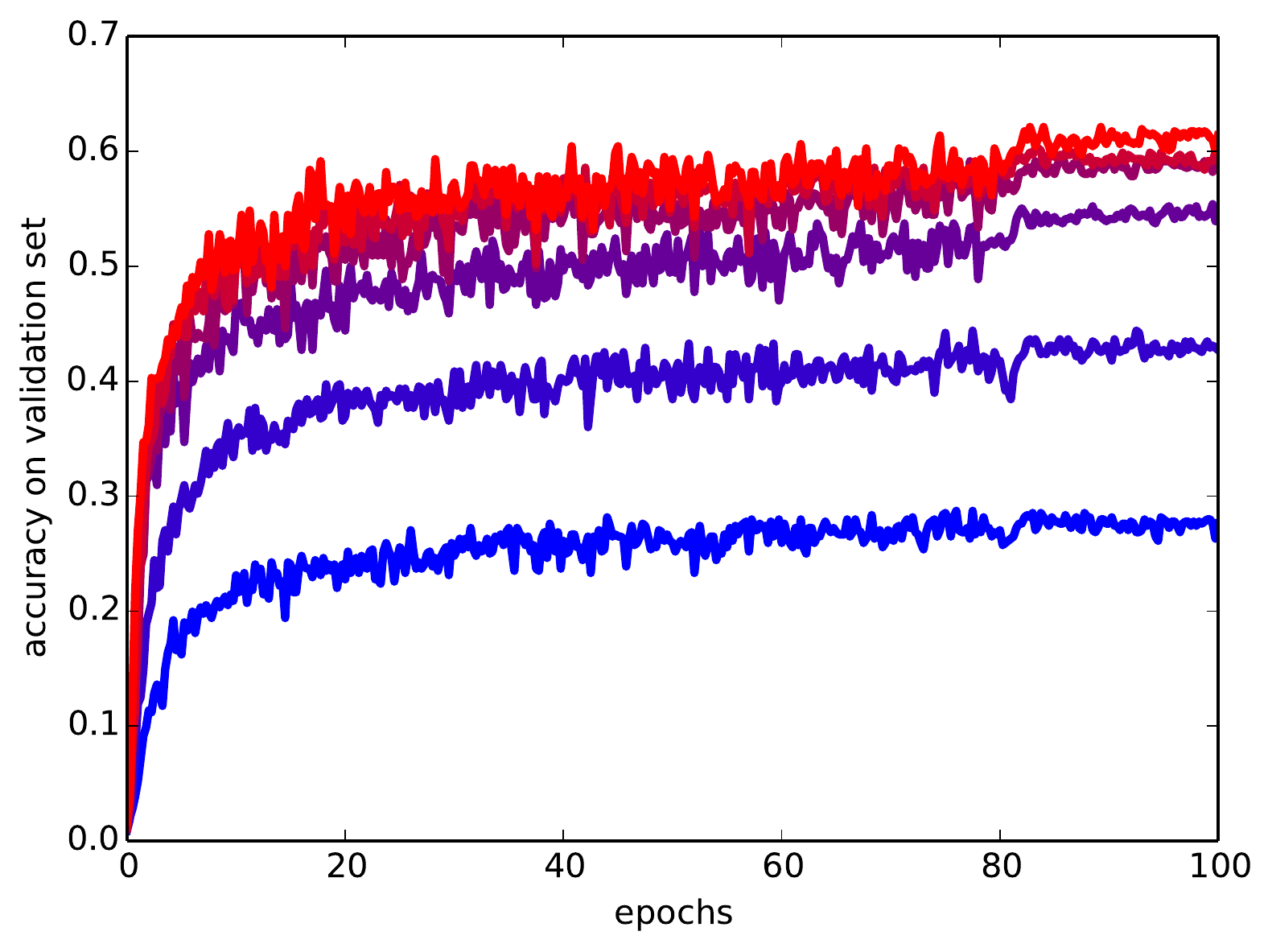}  %
        \caption{Convergence during learning an Impatient AlexNet trained on MIT-67 with (right) and without (left) batch normalization:
            Different colors indicate individual early prediction layers and it can be clearly seen that batch normalization significantly improves
            stability during training.
        }
        \label{fig:experiments:batchNormPlot}
    \end{figure}

    Therefore, we made use of batch normalization \cite{ioffe2015batch} which rectifies the covariate shift in the input data distribution of each convolution layer.
    This technique allows for training with much higher learning rates ensuring faster convergence and in our case convergence at all.
    In \figurename~\ref{fig:experiments:batchNormPlot} (left), an example of optimizing an Impatient AlexNet is shown where the validation accuracy for early prediction layers saturates very slow at a low value caused by a highly decreased learning rate of $10^{-4}$.
    Even no convergence is achieved for very early layers after running 2000 epochs of training.
    In contrast, adding batch normalization (right-hand side) allows for a $100\times$ higher learning rate resulting in very fast convergence at a high level of validation accuracy for all prediction layers.
    
    \myparagraph{Evaluation of early prediction architectures}
    
\begin{figure}[t]
        \centering
        \includegraphics[height=0.24\linewidth]{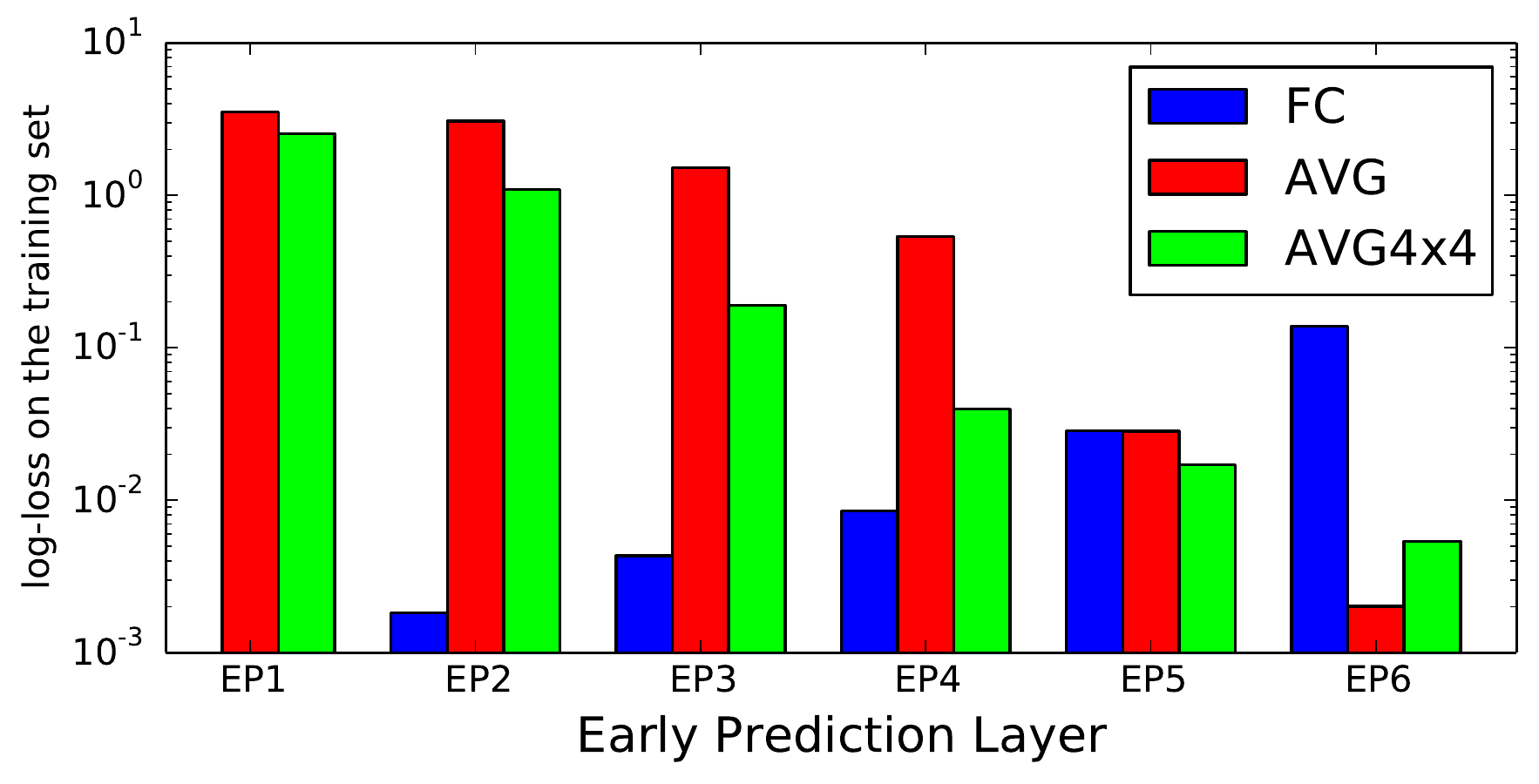} 
        \includegraphics[height=0.24\linewidth]{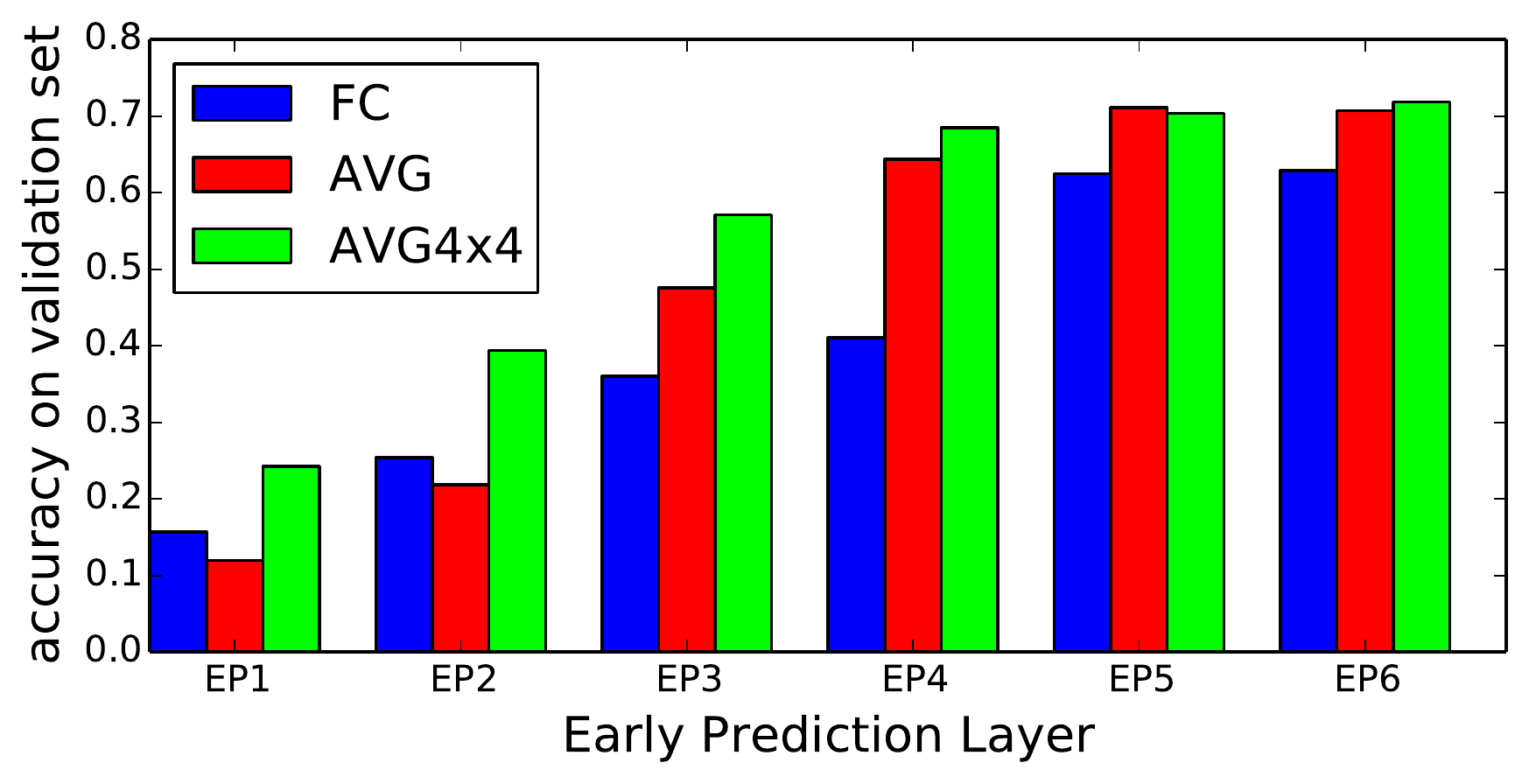} 
        \caption{Comparison of different early prediction architectures of an Impatient VGG19 trained on MIT-67.
            Replacing fully-connected layers (FC) by spatial average pooling (AVG \& AVG4x4) reduces the effect of overfitting resulting in higher validation accuracy.
        }
        \label{fig:experiments:EPs}
    \end{figure}

    As presented in \sectionname~\ref{sec:impatientcnns}, several architectures are possible for early prediction.
    The straightforward approach of connecting FC layers directly to each convolutional layer leads to a huge amount of additional parameters to be optimized.
    These layers are prone to overfitting.
    This can be seen in the learning statistics for MIT67 with a VGG19 base architecture shown in \figurename~\ref{fig:experiments:EPs}.
    The training loss is near zero together with a moderate validation accuracy for early layers.
    We also experimented with multiple FC layers.
    However, learning of these architectures failed to converge in all cases independently from the choice of hyperparameters.
    By applying spatial pooling layers, validation accuracy is substantially improved, 
    which can be seen in \figurename~\ref{fig:experiments:EPs} (AVG and AVG4x4).
    Especially AVG4x4 provides rough spatial information which helps to improve performance even further. Therefore, we use this architecture in the following experiments.
    
    In the last two columns of \tablename~\ref{tab:experiments:baselinesvgg}, average computation times according to the particular weighting schemes and budget distributions are presented for a single image.
    If inference is performed up to a particular prediction layer known in advance, previous prediction layers do not have to be assessed
    and we achieve low prediction times $t_{\text{B}}$ without additional overhead.
    Interruptable prediction in anytime scenario A ($t_{\text{A}}$) requires inference of all intermediate prediction layers caused by the potential sudden interruption.
    In the worst case, \ie\ the forward pass includes all prediction layers, average computation time increases compared to the 
    scenario with a-priori given budgets.
    All experiments were performed on an NVIDIA GeForce GTX 970 GPU.

    \myparagraph{Does joint training of EP layers help?}
    The most interesting question, however, is whether our joint training scheme motivated in \sectionname~\ref{sec:anytimepred} 
    provides superior results compared to learning predictors independently.
    To answer this question, we compared our approach with different baselines that learn several SVM classifiers based on extracted CNN features \cite{donahue2013decaf} at each early prediction layer.
    We optimize SVM hyperparameters on the validation set to allow fair comparison.
    The underlying networks, on the contrary, differ in the sense that we made use of an original CNN pre-trained on ImageNet and a pre-trained CNN fine-tuned on the current dataset.
    
    In \tablename~\ref{tab:experiments:baselinesvgg}, the evaluation for different time-budget distributions is presented where each result shows the expected
    accuracy according to the particular weighting scheme and budget distribution.
    It can be clearly seen that the original CNN (ORIG) without the adaptation to the current dataset performs worst.
    By applying fine-tuning (FT), however, accuracy can be noticeably increased for all early prediction SVMs.

    Our joint learning of the EP layers provides superior results in almost all scenarios.
    Especially in the case of small time budgets our method benefits from taking the budget distribution during learning into account 
    resulting in an improvement of almost 10\% on MIT-67 and 6\% on 15-Scenes for an Impatient VGG19 compared to the best performing baseline.
    For extreme weighting schemes with high priority on later predictions (\wscheme{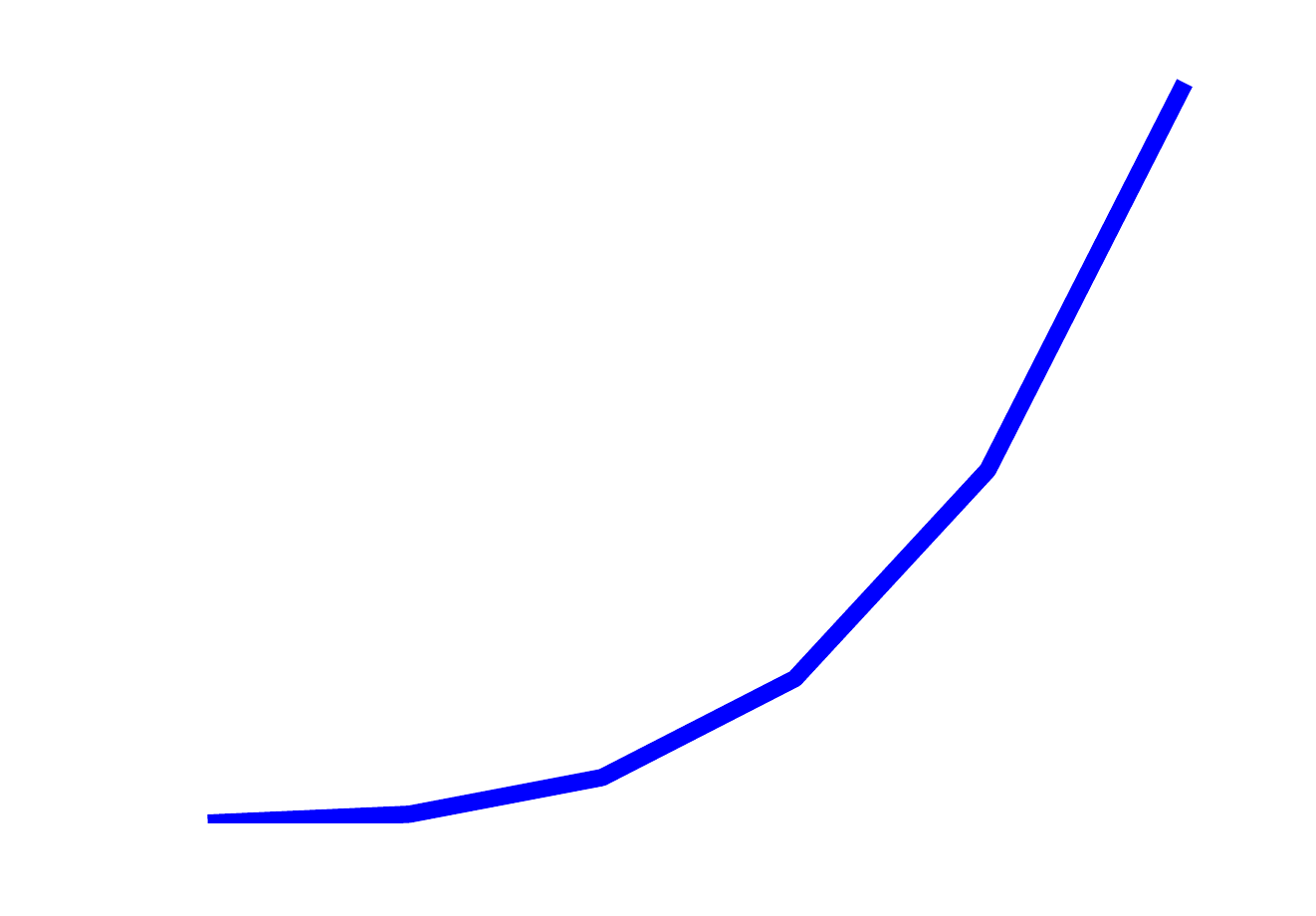}), fine-tuning of the original networks provides slightly better results compared to our approach.
    This is not surprising since in this case training is very similar to that of standard DNNs with only one final loss layer.
    
    In \tablename~\ref{tab:experiments:sota}, we compared our approach to state-of-the-art results for MIT-67 and 15-Scenes.
    Although the focus of this paper is rather on anytime capability while running the risk of dropping accuracy at final layers, we achieved superior results.
    It should be noted that only the last layer is used to obtain predictions, since we assume to have no budget restrictions.
    Especially for the jointly trained Impatient VGG19 on MIT-67, it was even possible to outperform the standard fine-tuned CNN, which supports the idea of ``deep supervision'' \cite{wang2015training}.

    \begin{table}[tb]
        \centering
        \resizebox{\linewidth}{!}{
        \setlength{\tabcolsep}{0.01\textwidth}
        \begin{tabular}{
                p{0.24\textwidth}  |
                C{0.12\textwidth}C{0.12\textwidth}C{0.12\textwidth}  |
                C{0.12\textwidth}C{0.12\textwidth}C{0.12\textwidth}  |
                C{0.12\textwidth}C{0.12\textwidth}
            }
            \toprule
            \thead{VGG19} & MIT-67 &&& 15-Scenes &&& & \\
            \thead{Budget Scheme}    & \thead{Orig} & \thead{FT} & \thead{Ours} 
            & \thead{Orig} & \thead{FT} & \thead{Ours} & $\varnothing t_{\text{B}}$ [ms] & $\varnothing t_{\text{A}}$ [ms]\\ 
            \midrule
            \wscheme{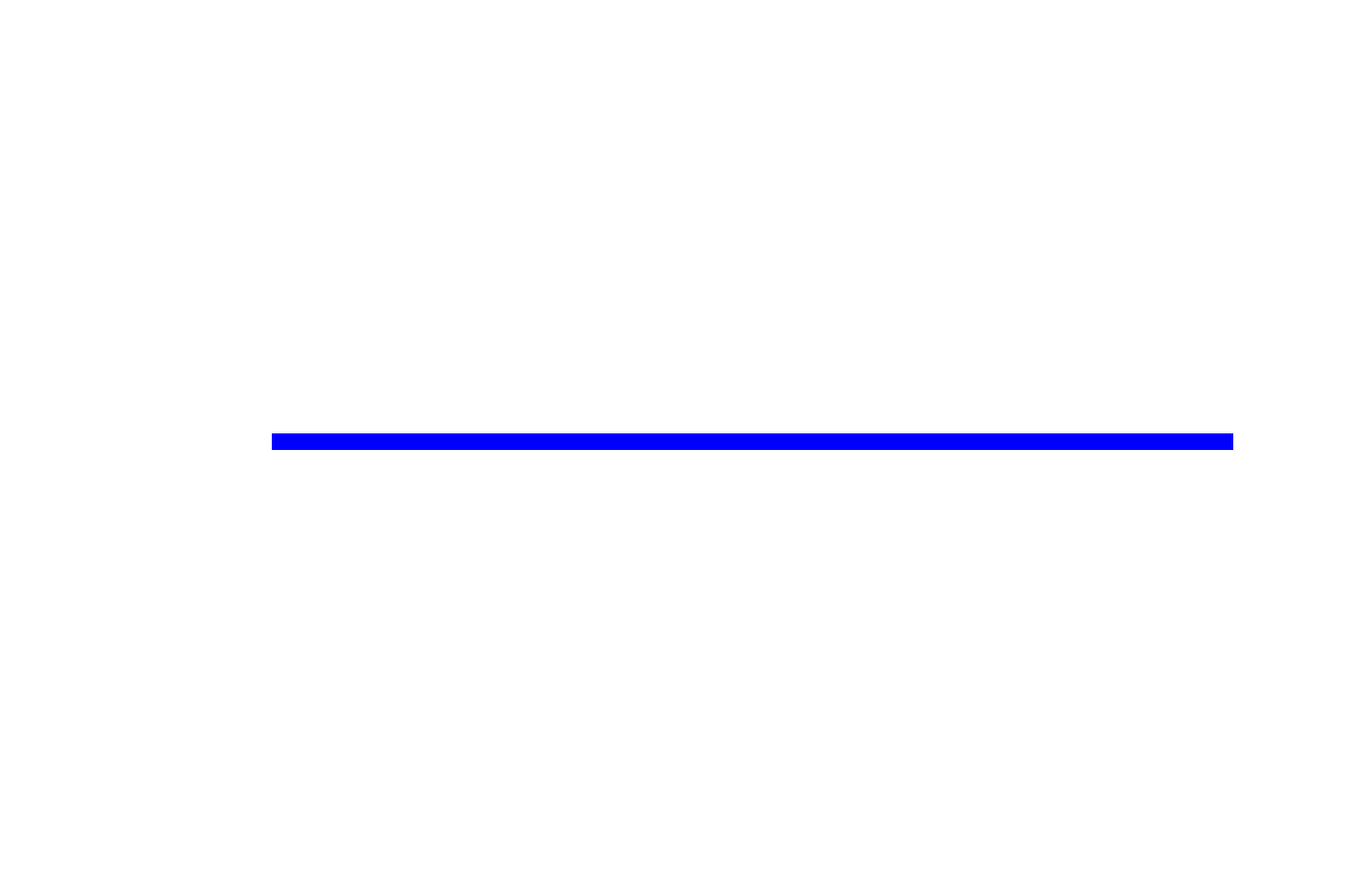}        & 46.65      & 48.07      & \bf{53.93}
                                & 83.37      & 84.28      & \bf{85.63}
                                & 1.11       & 1.19\\
            \wscheme{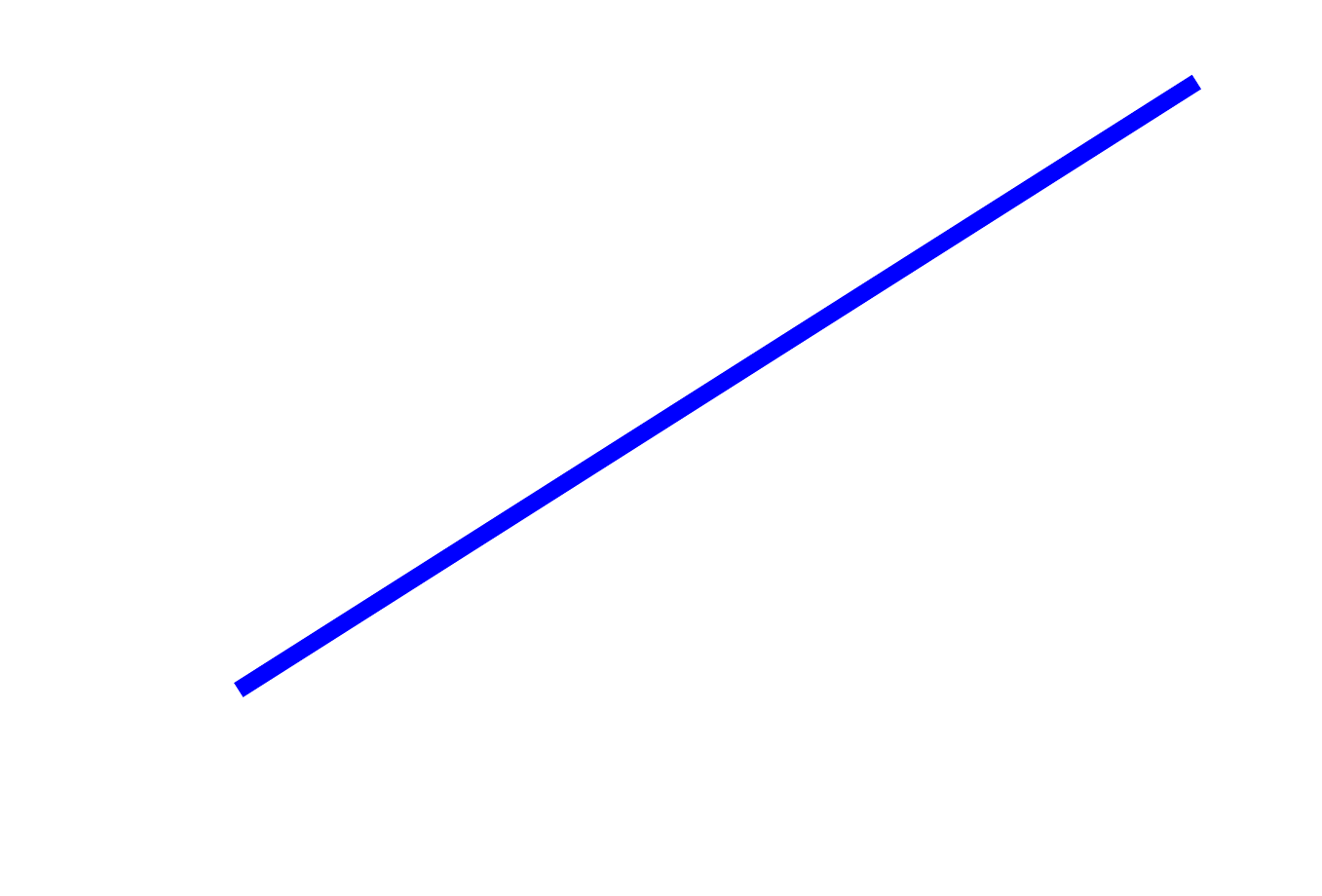}       & 54.19      & 56.52      & \bf{60.55}
                                & 85.87      & 87.47      & \bf{88.02}
                                & 1.37       & 1.47\\
            \wscheme{poly}      & 62.82      & 67.07      & \bf{69.66}
                                & 88.71      & \bf{91.71} & 90.88
                                & 1.72       & 1.84\\
            \wscheme{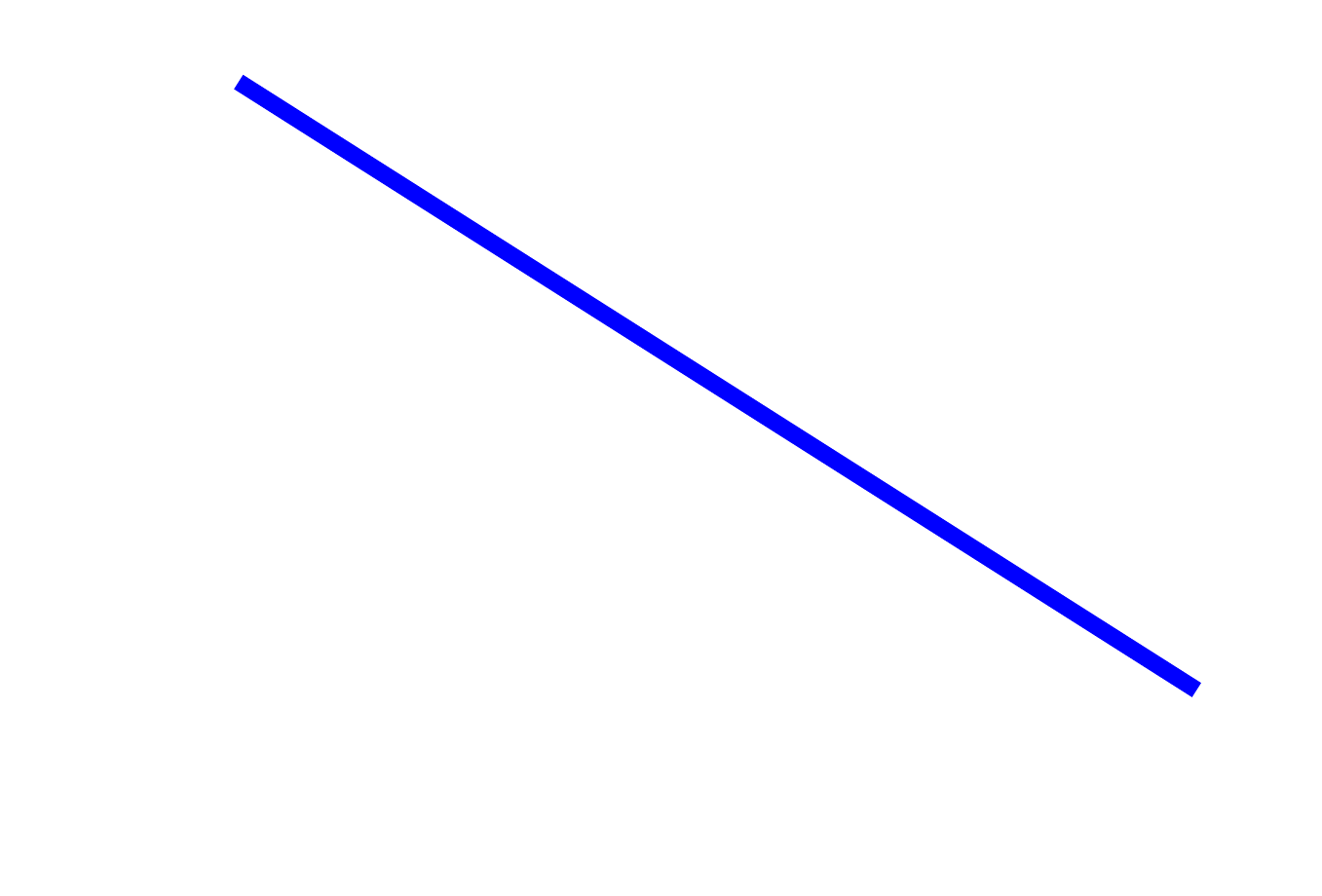}      & 37.25      & 37.71      & \bf{45.62}
                                & 77.56      & 77.73      & \bf{80.87}
                                & 0.82       & 0.86\\
            \wscheme{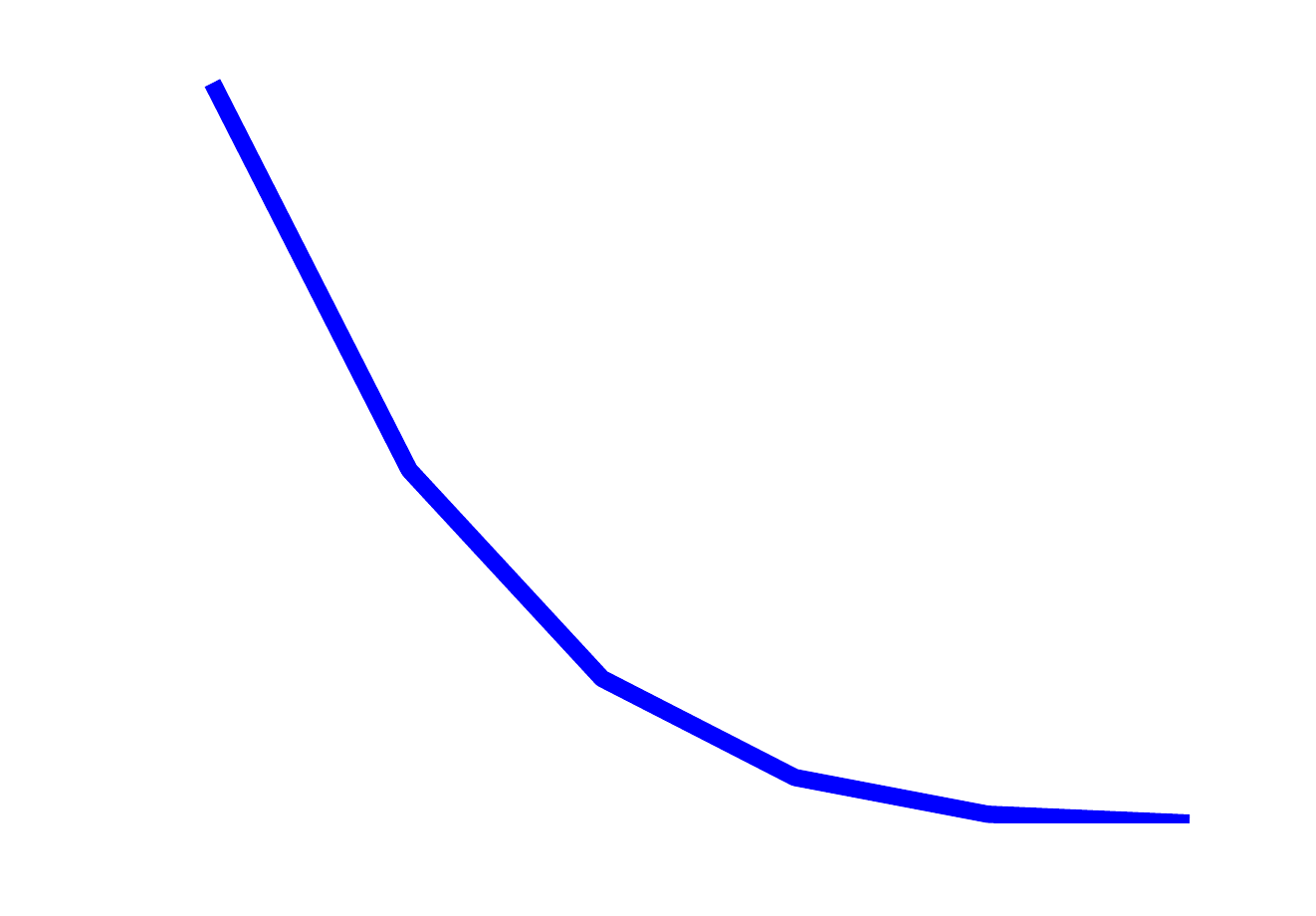}     & 25.63      & 25.65      & \bf{35.11}
                                & 70.14      & 69.85      & \bf{75.93}
                                & 0.50       & 0.51\\
            \wscheme{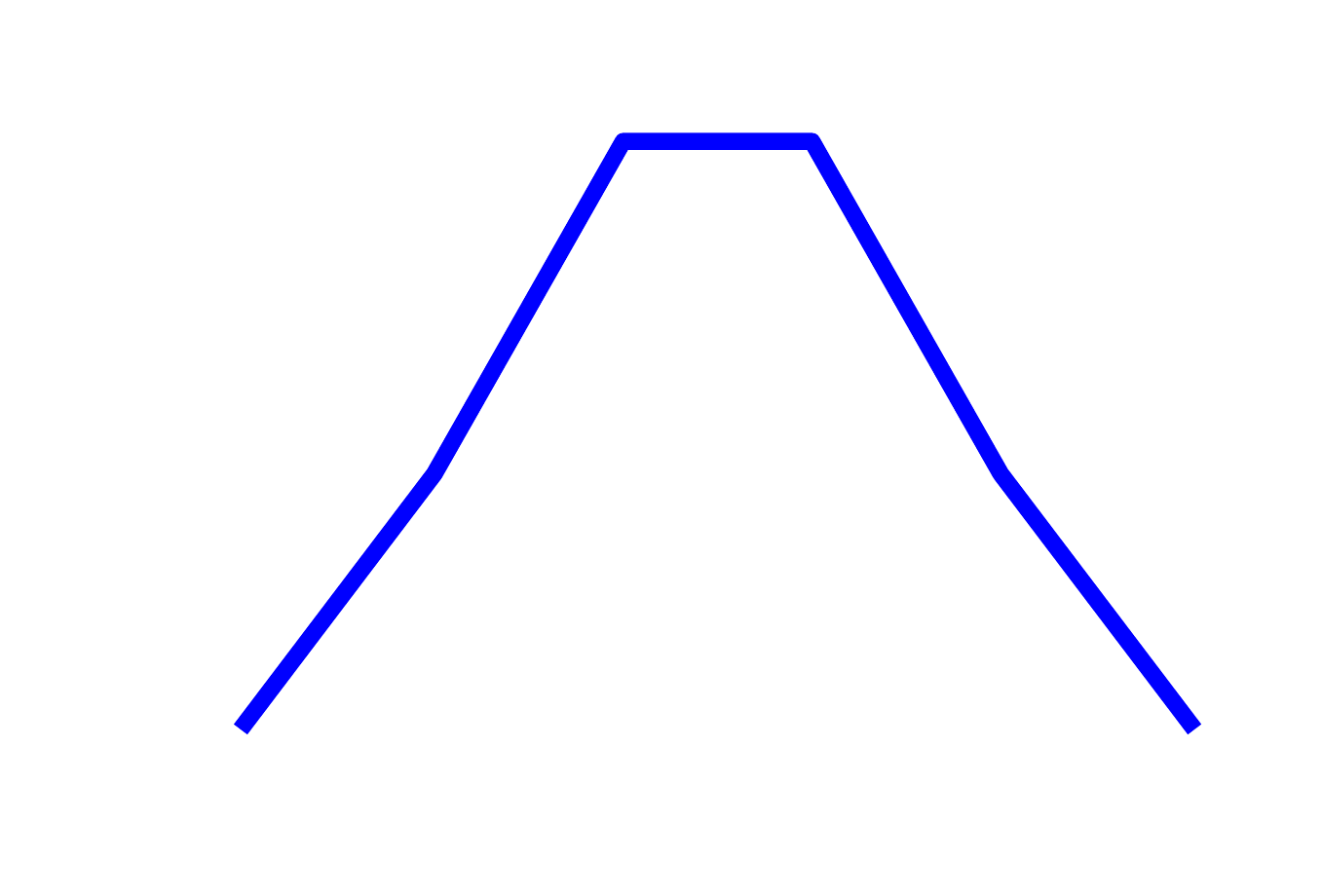}      & 47.53      & 47.90      & \bf{55.38}
                                & 84.46      & 84.74      & \bf{86.67}
                                & 1.07      & 1.15\\
            \bottomrule
        \end{tabular}
        }
        \resizebox{\linewidth}{!}{
        \setlength{\tabcolsep}{0.01\textwidth}
        \begin{tabular}{
                p{0.24\textwidth}  |
                C{0.12\textwidth}C{0.12\textwidth}C{0.12\textwidth}  |
                C{0.12\textwidth}C{0.12\textwidth}C{0.12\textwidth}  |
                C{0.12\textwidth}C{0.12\textwidth}
            }
            \toprule
            \thead{AlexNet} & MIT-67 &&& 15-Scenes &&& & \\
            \thead{Budget Scheme}    & \thead{Orig} & \thead{FT} & \thead{Ours} 
            & \thead{Orig} & \thead{FT} & \thead{Ours} & $\varnothing t_{\text{B}}$ [ms] & $\varnothing t_{\text{A}}$ [ms]\\ 
            \midrule
            \wscheme{eq}        & 41.75      & 46.19      & \bf{48.40}
                                & 82.56      & 84.28      & \bf{85.11}
                                & 0.68       & 0.75\\
            \wscheme{lin}       & 45.19      & 50.96      & \bf{52.13}
                                & 83.73      & \bf{86.19} & 85.94
                                & 0.79       & 0.89\\
            \wscheme{poly}      & 48.50      & \bf{56.29} & 55.76
                                & 85.56      & \bf{88.98} & 87.38
                                & 0.96       & 1.09\\
            \wscheme{ilin}      & 36.64      & 39.59      & \bf{42.91}
                                & 78.10      & 79.03      & \bf{81.87}
                                & 0.54       & 0.59\\
            \wscheme{ipoly}     & 28.69      & 30.17      & \bf{36.14}
                                & 72.38      & 72.48      & \bf{77.85}
                                & 0.40       & 0.42\\
            \wscheme{norm}      & 43.97      & 47.80      & \bf{49.93}
                                & 83.25      & 84.82      & \bf{84.89}
                                & 0.65       & 0.72\\
            \bottomrule
        \end{tabular}
        }
        \caption{
            Comparison of Impatient AlexNet (top) and VGG19 (bottom) CNNs with several baselines.
            Performance is measured by expected accuracy in $\%$ based on the particular budget distribution.
            \label{tab:experiments:baselinesvgg}
        }

    \end{table}
            
    \begin{table}
        \centering
        \begin{tabular}{lcccccc}
            \toprule
            Dataset & Orig & FT & Ours (eq) & Ours (poly) & PlacesCNN~\cite{zhou2014learning} & \cite{liu2015treasure}$^*$\\
            \midrule
            MIT-67 & $65.0\%$ & $71.04\%$ & $67.23\%$ & $\mathbf{71.71\%}$ & $68.24\%$ & $71.5\%$\\
            15-Scenes & $88.30\%$ & $\mathbf{92.83\%}$ & $92.13\%$ & $91.45\%$ & $90.19\%$ & -\\ 
            \bottomrule
        \end{tabular}
        \caption{How good are our VGG19 Impatient Networks when there are no budget restrictions during testing? The table shows the accuracy of the last prediction layer also compared to state-of-the-art results. $^*$ The method of \cite{liu2015treasure} requires more than $4s$ per image.}
            \label{tab:experiments:sota}
    \end{table}

    \myparagraph{Cascaded prediction}
    
    Apart from both scenarios presented in \figurename~\ref{fig:teaser}, efficient classification constitutes another interesting application of our approach.
    The task here is--for a given set of examples--to reach a desired accuracy within a minimal but not fixed amount of time.
    In particular, interrupting the network at a certain depth might already provide the correct decision which renders further computation unnecessary.
    To implement the idea of efficient inference, an adequate stopping criterion has to be defined.
    Since each early prediction layer provides probabilistic outputs, we applied uncertainty-based decision making by calculating the 
    ratio between the two highest class probabilities, which is known as 1-vs-2 strategy~\cite{joshi2009multi}. 
    If the current prediction of class probabilities is characterized by a high ratio, inference can be interrupted.
    
\begin{figure}[tb]
        \centering
        \includegraphics[height=0.205\linewidth]{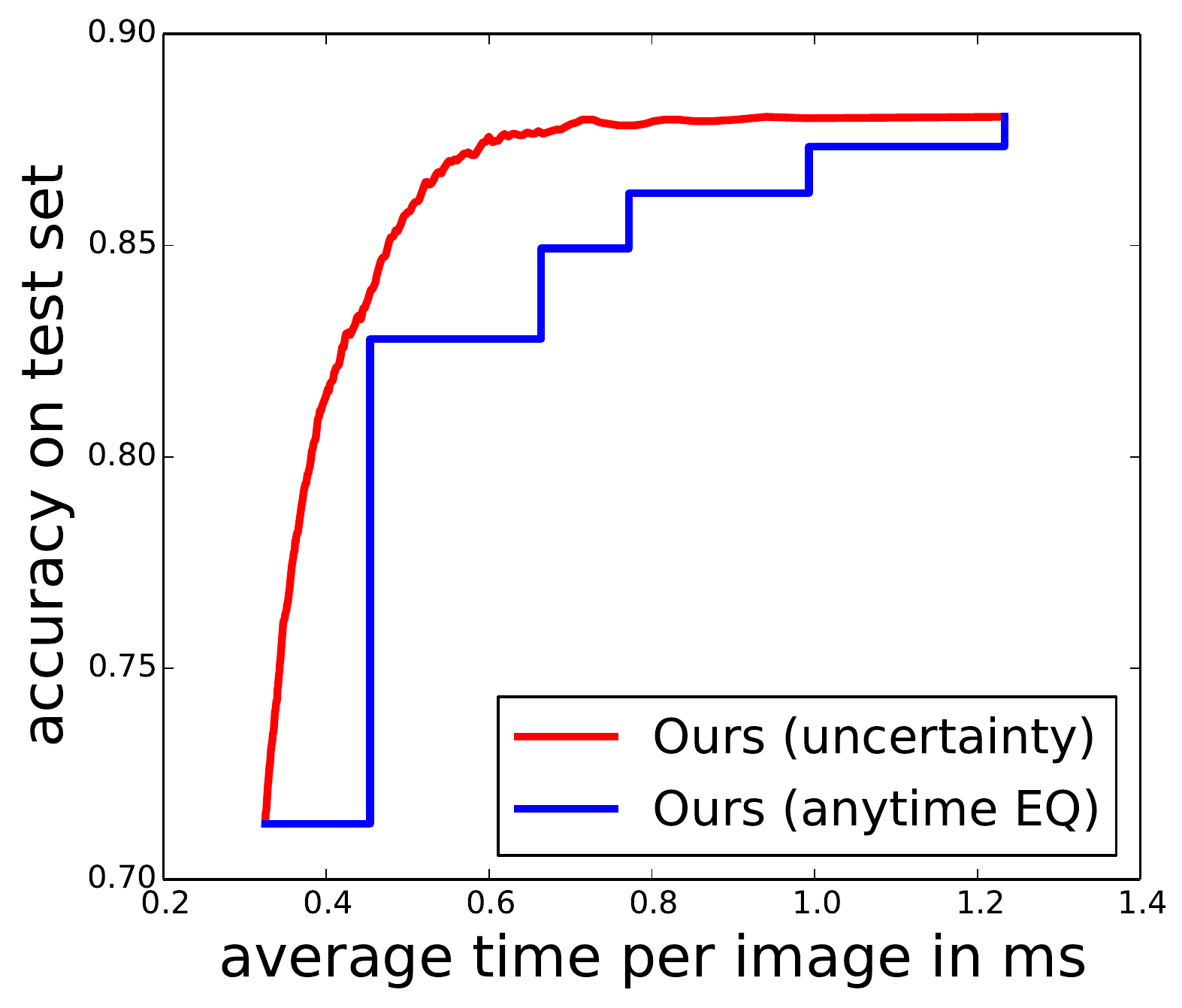}
        \includegraphics[height=0.205\linewidth]{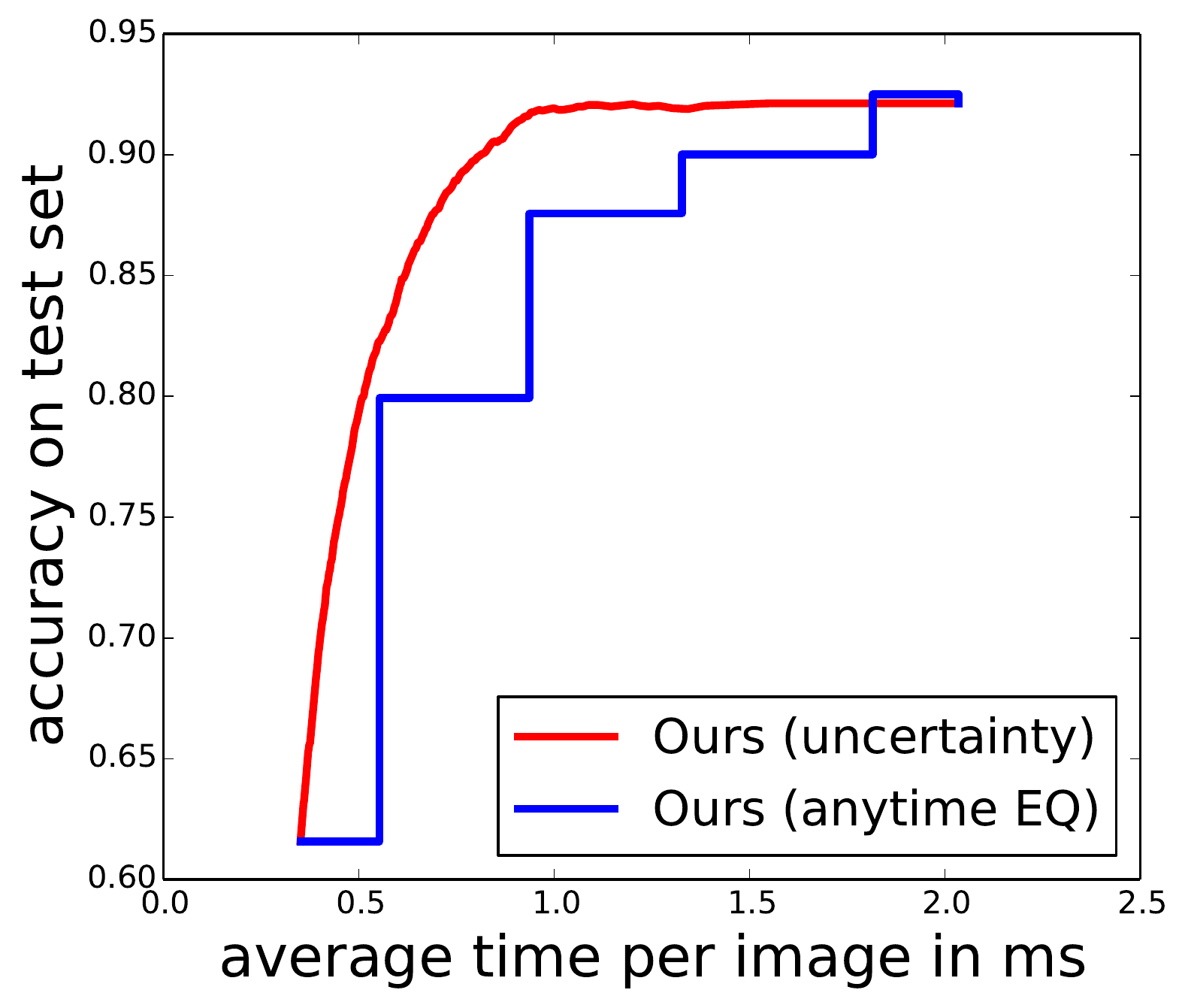} 
        \includegraphics[height=0.205\linewidth]{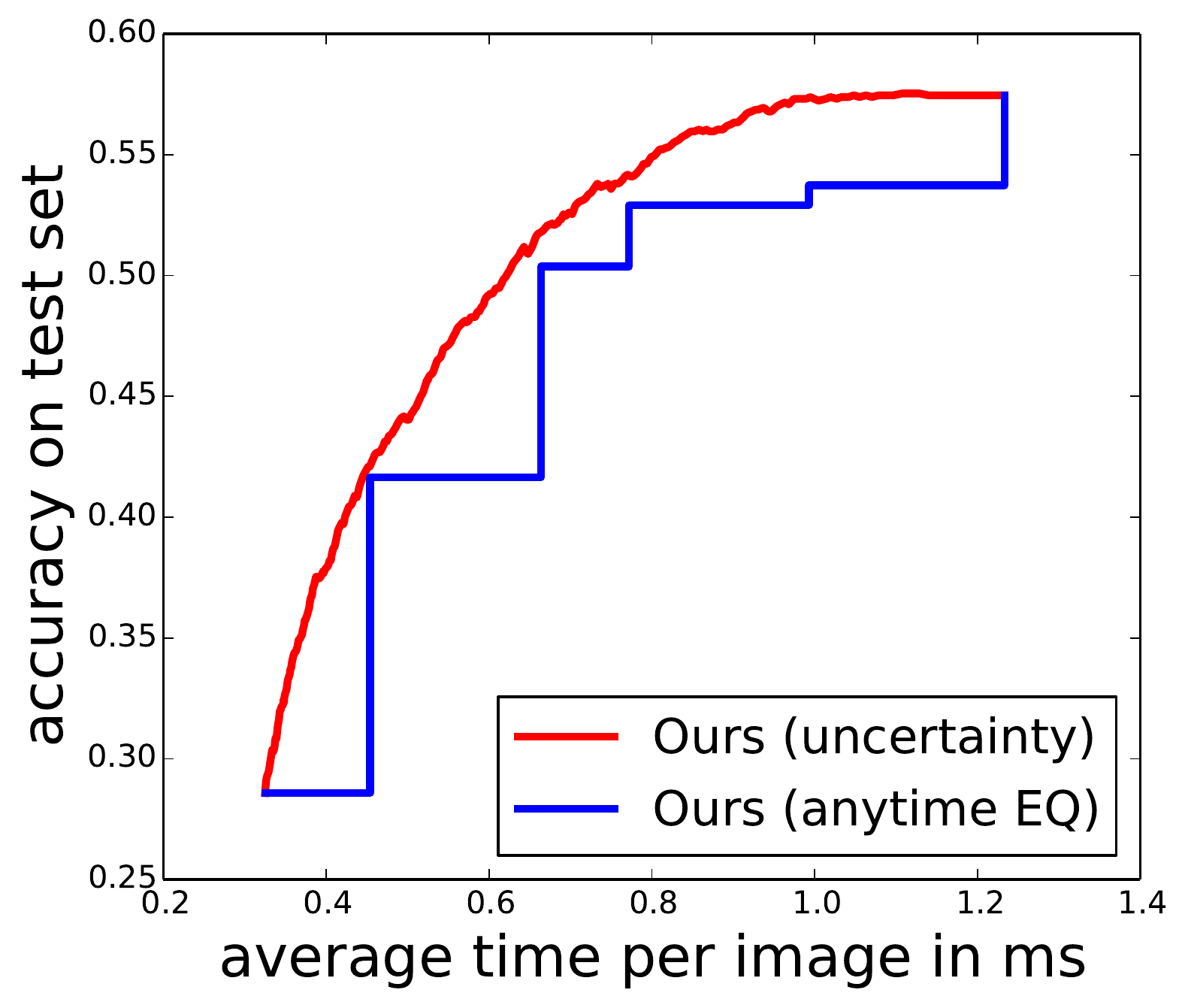}
        \includegraphics[height=0.205\linewidth]{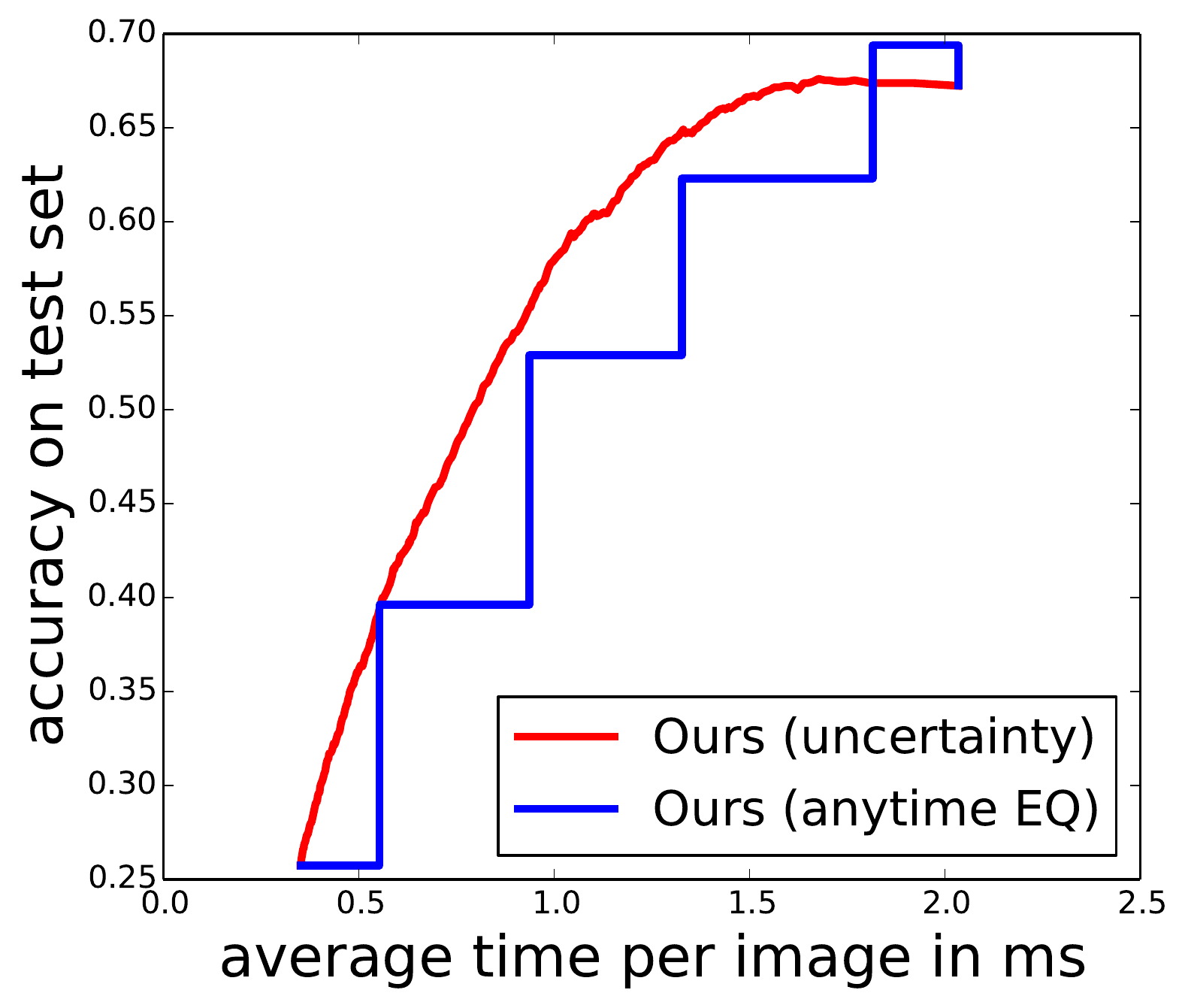} 
        \caption{
            Evaluation of uncertainty-based predictions compared to early layer predictions.
            From left to right: Impatient AlexNet on 15-Scenes, Impatient VGG19 on 15-Scenes, Impatient AlexNet on MIT-67, and Impatient VGG19 on MIT-67.
        }
        \label{fig:experiments:entropy}
    \end{figure}

    The analysis of the proposed criterion can be seen in \figurename~\ref{fig:experiments:entropy} showing time-accuracy plots.
    Thereby, one point on the red graph is obtained by a fixed ratio threshold which determines whether an early layer prediction already reaches sufficient certainty and thus provides the final decision.
    The blue graph, however, represents classification results of each early prediction layer itself, \ie, the final decision is made at always the same depth, independently of the underlying ratio.
    As can be seen, by using uncertainty-based predictions, accuracy can be increased substantially in a lot of cases with the same computational efforts.
    For example, by interrupting the AlexNet network at the fifth prediction layer consistently takes $\sim$ 1 ms per image for MIT-67 (second-last plot in \figurename~\ref{fig:experiments:entropy}).
    In contrast, using the proposed criterion, accuracy can be increased from 53\% up to 57\% while still requiring exactly the same computation time on average.
    An entropy-based criterion achieved inferior performance in our experiments.
    
    \myparagraph{Qualitative results}
    
\begin{figure}[tb]
        \centering
        \includegraphics[width=0.99\linewidth]{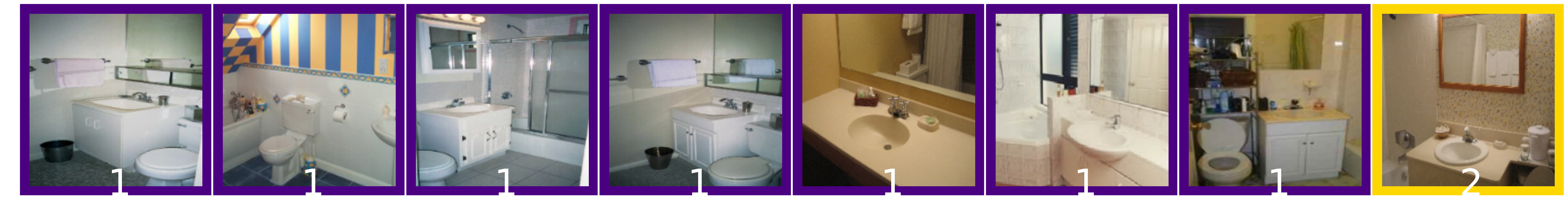} %
        \includegraphics[width=0.99\linewidth]{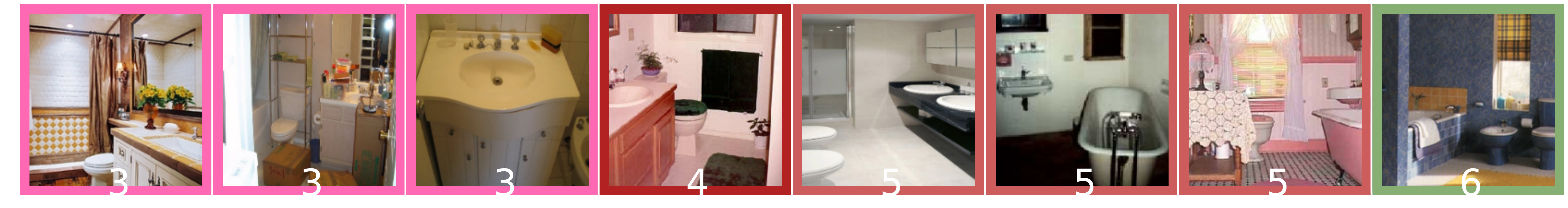} %
        \caption{
            Images of the MIT-67 first correctly classified as ``bathroom'' at different early prediction layers of an Impatient VGG19 CNN.
            The position of the layers is highlighted as a number and a uniquely colored border.
        }
        \label{fig:experiments:qualitative}
    \end{figure}

    In \figurename~\ref{fig:experiments:qualitative}, qualitative results for the task of scene recognition (class ``bathroom'' from MIT-67) are shown.
    Different numbers in each image indicate the early prediction layer in which the particular example was first correctly classified.
    It can be clearly seen that the examples already decided at EP1 are white colored bathrooms with clearly visible toilet bowl, shower, and sink.
    With increasing complexity of the scene, layer depth increases as well to provide correct decisions.
    For example, the right most images in the second row of \figurename~\ref{fig:experiments:qualitative} shows extraordinary bathrooms of unusual colored walls and furnishings increasing the likelihood of confusion with other classes, \eg\ children room.
    
\section{Conclusions}
\label{sec:conclusions}

    In this paper, we presented impatient deep neural networks that tackle the problem of classification with dynamic time budgets during application.
    Compared to standard DNNs which suffer from a high computational demand during inference, we showed that our approach allows for anytime prediction, \ie a possible interruption at multiple stages while still providing output estimates which renders our method suitable even for real-time applications.
    We presented a novel general framework of learning dynamic budget predictors based on risk minimization, which we adapted directly to state-of-the-art convolutional neural network architectures by branching additional early prediction layers with weighted losses.
    Based on a set of object classification datasets and architectures, we showed that our approach provides superior results for different time budget distributions.
    Furthermore, we developed an uncertainty-based prediction framework allowing for reducing computational costs while still providing the same accuracy.
    
\bibliography{paper}

\end{document}